\documentclass[10pt,doublecolumn]{IEEEtran}
\usepackage[mathscr]{eucal}
\usepackage[cmex10]{amsmath}
\usepackage{epsfig,epsf,psfrag}
\usepackage{amssymb,amsmath,amsthm,amsfonts,latexsym}
\usepackage{amsmath,graphicx,bm,xcolor,url,overpic}
\usepackage{array}
\usepackage{verbatim}
\usepackage{bm}
\usepackage{algorithm}
\usepackage{verbatim}
\usepackage{textcomp}
\usepackage{mathrsfs}
\usepackage{epstopdf}

\newcommand{\openone}{\leavevmode\hbox{\small1\normalsize\kern-.33em1}}

\catcode`~=11 \def\UrlSpecials{\do\~{\kern -.15em\lower .7ex\hbox{~}\kern .04em}} \catcode`~=13 

\allowdisplaybreaks[4]

\newcommand{\nn}{\nonumber}

\newcommand{\calF}{\mathcal{F}}

\newcommand{\calM}{\mathcal{M}}
\newcommand{\calN}{\mathcal{N}}

\newcommand{\calP}{\mathcal{P}}

\newcommand{\calR}{\mathcal{R}}
\newcommand{\calS}{\mathcal{S}}
\newcommand{\calT}{\mathcal{T}}

\newcommand{\calX}{\mathcal{X}}
\newcommand{\calY}{\mathcal{Y}}

\newcommand{\bP}{\mathbf{P}}

\newcommand{\by}{\mathbf{y}}
\newcommand{\bY}{\mathbf{Y}}

\newcommand{\rmseq}{\mathrm{seq}}
\newcommand{\rmtp}{\mathrm{tp}}

\newcommand{\rmH}{\mathrm{H}}

\newcommand{\rmP}{\mathrm{P}}

\newcommand{\rmr}{\mathrm{r}}

\newcommand{\bbE}{\mathsf{E}}

\newcommand{\bbN}{\mathbb{N}}

\newcommand{\bbP}{\mathbb{P}}

\newcommand{\bbR}{\mathbb{R}}

\DeclareMathAlphabet{\mathbsf}{OT1}{cmss}{bx}{n}
\DeclareMathAlphabet{\mathssf}{OT1}{cmss}{m}{sl}

\DeclareSymbolFont{bsfletters}{OT1}{cmss}{bx}{n}  
\DeclareSymbolFont{ssfletters}{OT1}{cmss}{m}{n}
\DeclareMathSymbol{\bsfGamma}{0}{bsfletters}{'000}
\DeclareMathSymbol{\ssfGamma}{0}{ssfletters}{'000}
\DeclareMathSymbol{\bsfDelta}{0}{bsfletters}{'001}
\DeclareMathSymbol{\ssfDelta}{0}{ssfletters}{'001}
\DeclareMathSymbol{\bsfTheta}{0}{bsfletters}{'002}
\DeclareMathSymbol{\ssfTheta}{0}{ssfletters}{'002}
\DeclareMathSymbol{\bsfLambda}{0}{bsfletters}{'003}
\DeclareMathSymbol{\ssfLambda}{0}{ssfletters}{'003}
\DeclareMathSymbol{\bsfXi}{0}{bsfletters}{'004}
\DeclareMathSymbol{\ssfXi}{0}{ssfletters}{'004}
\DeclareMathSymbol{\bsfPi}{0}{bsfletters}{'005}
\DeclareMathSymbol{\ssfPi}{0}{ssfletters}{'005}
\DeclareMathSymbol{\bsfSigma}{0}{bsfletters}{'006}
\DeclareMathSymbol{\ssfSigma}{0}{ssfletters}{'006}
\DeclareMathSymbol{\bsfUpsilon}{0}{bsfletters}{'007}
\DeclareMathSymbol{\ssfUpsilon}{0}{ssfletters}{'007}
\DeclareMathSymbol{\bsfPhi}{0}{bsfletters}{'010}
\DeclareMathSymbol{\ssfPhi}{0}{ssfletters}{'010}
\DeclareMathSymbol{\bsfPsi}{0}{bsfletters}{'011}
\DeclareMathSymbol{\ssfPsi}{0}{ssfletters}{'011}
\DeclareMathSymbol{\bsfOmega}{0}{bsfletters}{'012}
\DeclareMathSymbol{\ssfOmega}{0}{ssfletters}{'012}

\newcommand{\tilx}{\tilde{x}}

\newcommand{\tily}{\tilde{y}}

\newcommand{\barg}{\bar{g}}

\newcommand{\barD}{\bar{D}}

\DeclareMathOperator*{\argmin}{arg\,min}

\newtheorem{theorem}{Theorem} 
\newtheorem{lemma}{Lemma}

\usepackage{graphicx,amssymb,amstext,amsmath}
\usepackage{makecell}
\usepackage{amsthm}
\usepackage{color}
\usepackage{float}
\usepackage{subfigure}
\usepackage{algpseudocode}
\usepackage{multirow}
\usepackage[switch,pagewise]{lineno}
\usepackage{enumerate}
\usepackage[colorlinks = true,
linkcolor = blue,
urlcolor  = blue,
citecolor = red,
anchorcolor = green,]{hyperref}
\usepackage{color}

\definecolor{Dyellow}{RGB}{254,152,0}
\definecolor{Dgreen}{RGB}{0,176,80}

\begin{document}
\title{Exponentially Consistent Statistical Classification of Continuous Sequences with Distribution Uncertainty}

\author{\IEEEauthorblockN{Lina Zhu and Lin Zhou}\\
\thanks{L. Zhu is with Space Information Research Institute, Hangzhou Dianzi University, Hangzhou, Zhejiang, China, 310018 (Email: zhulina@hdu.edu.cn). L. Zhou is with the School of Cyber Science and Technology, Beihang University, Beijing 100191, China (Email: lzhou@buaa.edu.cn).}
}

\maketitle
\flushbottom

\begin{abstract}
In multiple classification, one aims to determine whether a testing sequence is
generated from the same distribution as one of the $M$ training sequences or not. Unlike most of existing studies that focus on discrete-valued sequences with perfect distribution match, we study multiple classification for continuous sequences with \emph{distribution uncertainty}, where the generating distributions of the testing and training sequences deviate even under the true hypothesis. In particular, we propose distribution free tests and prove that the error probabilities of our tests decay exponentially fast for three different test designs: fixed-length, sequential, and two-phase tests. We first consider the simple case without the null hypothesis, where the testing sequence is known to be generated from a distribution close to the generating distribution of one of the training sequences. Subsequently, we generalize our results to a more general case with the null hypothesis by allowing the testing sequence to be generated from a distribution that is vastly different from the generating distributions of all training sequences.
\end{abstract}


\section{Introduction}
\label{sec:intro}

In binary hypothesis testing, one is given two known distributions, $P_1$ and $P_2$, and an observed testing sequence $Y^n=(Y_1,\ldots,Y_n)$. The task is to decide whether $Y^n$ is generated i.i.d. from $P_1$ or $P_2$. 
There are two types of error events for the binary hypothesis testing: type-I error and type-II error events. A type-I error event occurs if $Y^n$ is claimed to be generated from $P_2$ while $Y^n$ is actually generated from $P_1$ and a type-II error event is defined similarly. The tradeoff between type-I and type-II error probabilities for fixed-length binary hypothesis tests has been characterized by the Chernoff-Stein lemma~\cite{Chernoff1952AMO} in the Neyman-Pearson setting and by Blahut~\cite{Blahut1974HypothesisTA} for the Bayesian setting. Subsequently, Wald \cite{wald1948optimum} showed that the tradeoff between error probabilities can be resolved by sequential tests. However, the superior performance of an optimal sequential test is achieved at the cost of high design complexity. To balance the design complexity and performance between the fixed-length and the sequential tests for binary hypothesis testing, Lalitha and Javidi~\cite{AFL} proposed a two-phase test and demonstrated that the test could achieve error exponents close to the sequential test with design complexity similar to a fixed-length test. 

In many real world machine leaning applications such as computer vision and image classification, the assumption of knowing the generating distributions in binary hypothesis testing is impractical ~\cite{ziv1988classification,Asymptotically_optimal_classification}. To model this case, binary classification is proposed, where one is given a testing sequence $Y^n$ and two training sequences $(X_1^n,X_2^n)$, where for each $i\in\{1,2\}$, $X_i^n$ is generated i.i.d. from an unknown distribution $P_i$. The task is to determine whether $Y^n$ is generated i.i.d. from $P_1$ or $P_2$. Binary classification can be further generalized to multiple classification, where the number of training sequences is $M\geq 2$. Gutman~\cite{Asymptotically_optimal_classification} characterized the performance of optimal fixed-length tests using empirically observed statistics. The performance of sequential tests was characterized by Haghifam, Tan and Khsiti \cite{Sequential_classification} and by Hsu, Li and Wang \cite{hsu2022universal}. The two-phase tests were proposed and analyzed by Diao, Zhou and Bai \cite{bai2022achievable,diao2023achievable}.

The above studies assume that the generating distributions of training and testing sequences are perfectly matched, i.e., the testing sequence is generated from the same distribution as one of the training sequences. However, in many practical applications, there exists \emph{distribution uncertainty}, where the generating distribution of the testing sequence deviates from the generating distributions of training sequences. In this case, the testing sequence is claimed to be generated from the same distribution as one training sequence if their generating distributions are close to each other under a certain distribution distance measure. To close the research gap, Hsu \emph{et al.}~\cite{binaryI-Hsiang} studied binary statistical classification with mismatched empirically observed statistics, and analyzed the error exponents of optimal tests in both Stein's regime and Chernoff's regime.

Till now, all above studies consider \emph{discrete-valued} sequences and the corresponding results for \emph{continuous} sequences are limited.  The only known result is by Li \emph{et al.}~\cite{li2018nonparametric} for fixed-length multiple classification, where the authors introduced a fixed-length test leveraging the Maximum Mean Discrepancy (MMD) metric, incorporating distributional uncertainty, and conducted an analysis of its achievable error exponents. However, the results in~\cite{li2018nonparametric} have several limitations. Firstly, the results were only established for fixed-length tests while both sequential and two-phase tests were not studied. Secondly, for fixed-length tests, \cite{li2018nonparametric} only considered the case where the lengths of training and testing sequences are the same and ignored the impact of the ratio between the lengths of training and testing sequences. Finally, \cite{li2018nonparametric} did not consider the more practical setting with the null hypothesis scenario, where the testing sequence is generated from a distribution that is vastly different from the generating distribution of any training sequence.

To resolve above issues, we propose distribution free tests and prove that error probabilities of our tests decay exponentially fast for three different test designs: fixed-length test, sequential test, and two-phase test. In a fixed-length test, one fixes the sample size of each training and testing sequences; in a sequential test, one takes a sample sequentially from the testing sequence and takes a multiplicative number of samples from each training sequence until a reliable decision can be made; in a two-phase test, one adapts the sample sizes of training and testing sequences between two fixed values. Furthermore, we also analyze the effect of the distribution uncertainty on the performance of our proposed tests.
Our main contributions are summarized in the next subsection.

\subsection{Main contributions}

We first consider the simple case without the null hypothesis, i.e., the testing sequence is known to be generated from a distribution close to the generating distribution of one of the training sequences. In this case, each hypothesis specifies a possible distribution cluster that generating distribution of the testing sequence belongs to. A misclassification error event occurs if the test makes an error under any hypothesis. We propose three test designs: fixed-length, sequential, and two-phase tests, and prove that the misclassification probabilities of our tests decay exponentially fast. We show that the sequential test achieves larger exponents than the fixed-length test. Furthermore, analytically and numerically, we show that by changing design parameters, our two-phase test strikes a good balance between test design complexity and classification performance since it either reduces to a fixed-length test or approximately achieves the performance of a sequential test.

We next consider the more general case with the null hypothesis, where the testing sequence is allowed to be generated from an arbitrary distribution that is vastly different from the generating distribution of any training sequence.  In this case, we have both misclassification error and false alarm error events, where the latter event occurs if the null hypothesis is true but the test claims otherwise. In this case, we also propose three test designs and bound the exponential decay rates for misclassification and false alarm probabilities. Analogous to the simple case, we show that the two-phase test bridges over the fixed-length test and the sequential test by having performance close to the sequential test and having design complexity propositional to the fixed-length test. Furthermore, analytically and numerically, we show that there is a penalty of not knowing whether the null hypothesis is true.

\subsection{Organization of the Rest of the Paper}

The paper is organized as follows. Section \ref{Problem_formulation} covers notation, the multiple classification problem, and the MMD metric. In Section \ref{Main}, we present theoretical results for the simple case. In Section \ref{Main_un}, we generalize our results to the general case with the null hypothesis. Numerical examples are presented in Section \ref{simulation} to demonstrate our theoretical benchmarks. The paper concludes with Section \ref{sec:conc}, where we summarize our findings and propose avenues for future research. To maintain the clarity and flow of the main text, all proofs are relegated to the appendices.


\subsection*{Notation}

We denote the sets of real numbers, non-negative real numbers, and natural numbers by $\calR$, $\calR_+$, and $\bbN$, respectively. All logarithms are taken to the base $e$. Sets are indicated using calligraphic font (e.g., $\calX$). Random variables are represented by uppercase letters (e.g., $X$) and their realizations by lowercase letters (e.g.,  $x$). A random vector of length $n\in\bbN$ is denoted as $Y^n = (Y_1,\ldots,Y_n)$. The collection of all probability density functions (pdfs) on $\calR$ is represented by $\calP(\calR)$. For integers $(a,b)\in\bbN^2$, $[a:b]$ refers to the set of integers from $a$ to $b$, and $[a]$ denotes the set $[1:a]$.

\section{Problem Formulation}
\label{Problem_formulation}

\subsection{Simple Case}
In this subsection, we formulate the problem of statistical classification with multiple hypotheses and distribution uncertainty without the null hypothesis, i.e., the testing sequence is known to be generated from a distribution close to the generating distribution of one of the training sequences.

Given a distribution $P\in\calP(\calX)$, a positive real number $\delta\in\bbR_+$, and a distribution distance measure $d:\calP(\calX)^2\to\bbR_+$, define the following set of distributions
\begin{align}\label{MMDDistenceSet}
\calS_\delta(P) & :=\{Q\in\calP(\calX):~d(P,Q)\leq \delta\}.
\end{align}
Given any $(\delta,M)\in\bbR_+\times\bbN$ and any tuple of distributions $\bP:=(P_1,\ldots,P_M)\in\calP(\calX)^M$, we define the following two functions:
\begin{align}
&D_1(\delta,M)=\min_{i\in[M]}\min_{Q\in\bigcup_{j\in\calM_i}\calS_{\delta}(P_j)} \mathrm{MMD}^2(Q,P_i),\label{D1}\\
&D_2(\delta,M)=\max_{i\in[M]}\max_{Q\in\calS_{\delta}(P_i)} \mathrm{MMD}^2(Q,P_i),\label{D2}
\end{align}
where $\calM_i:=\{j\in[M]:~j\neq i\}$ denotes the set of all integers in $[M]$ except $i$. Note that here $D_1(\delta,M)$ is the minimal inter-cluster distance between different set of distributions, while $D_2(\delta,M)$ is the maximal intra-cluster distance of a distributions set. 

Fix three integers $(M,N,n)\in\bbN$. Let $\bP=(P_1,\ldots,P_M)\in\calP(\calX)^M$ be a tuple of $M$ probability distribution functions. In fixed-length statistical classification, one is given $M$ training sequences $\bY^N=\{Y_1^N,\ldots,Y_M^N\}\in (\calX^N)^M$ and a testing sequence $X^n$. For each $j\in[M]$, the training sequence $Y_j^N$ is generated i.i.d. from the generating distribution $P_j$. The testing sequence $X^n$ is generated from a distribution $Q$, which is close to one of the unknown generating distributions, i.e., $Q\in\calS_\delta(P_{i^*})$ for some $i^*\in[M]$. The task is to identify $i^*$ using the training and testing sequences. In order to make the problem meaningful, we should have  $D_1(\delta,M)>D_2(\delta,M)$ as the constraints of $\bP$ since otherwise, no reliable decision can be made.

We consider a potential sequential setting for sample collection, where both training sequences and testing sequences are collected in a streaming manner. To do so, we fix the ratio between the lengths of training and the testing sequence as a positive constant $\alpha\in\bbR_+$. In other words, for each $n\in\bbN$, if $n$ testing samples $X^n$ are given, then for each training sequence, $N=\alpha n$\footnote{We ignore the integer constraint for simplicity.} training samples are given. Under the above setting, our task is to design a test $\Phi=(\tau, \phi_\tau)$, which includes a potentially random stopping time $\tau\in\bbN$ and a mapping rule $\phi_\tau: \{X^\tau,\bY^{\alpha\tau}\}\to \{\rmH_1,\rmH_2,\ldots,\rmH_M\}$, to decide among the following $M$ hypotheses:
\begin{itemize}
\item $\rmH_i$,~$i\in[M]$: the testing sequence is generated i.i.d. from a distribution $Q$ that is close to the generating distribution of $i$-th training sequence, i.e., $Q\in \calS_\delta(P_i)$.
\end{itemize}
Note that the random stopping time $\tau$ is with respect to the filtration $\{\calF_n\}_{n\in\bbN}$, where $\calF_n$ is generated by $\sigma$-algebra $\sigma\{X^n, \bY_1^{n\alpha},\ldots \bY_M^{n\alpha}\}$ for each $n\in\bbN$.

Under any tuple of distributions $\bP\in \calP(\calX)^M$, to evaluate the performance of a test $\Phi=(\tau, \phi_\tau)$, for each $i\in[M]$, we consider the following misclassification error probability under hypothesis $\rmH_i$:
\begin{align}
\beta_i(\phi_\tau|\bP,\delta)& :=\max_{Q\in\calS_{\delta}(P_i)} \bbP_i\{\phi_\tau(X^\tau,\bY^{\alpha\tau})\neq\rmH_i\},\label{Error_define}
\end{align}
where under distribution $\bbP_i$, $X^\tau$ is generated i.i.d. from an unknown distribution $Q\in\calS_\delta(P_i)$. Note that $\beta_i(\phi_\tau|\bP,\delta)$ bounds the probability of the misclassification error event where the test outputs a hypothesis $\rmH_j$ with $j\neq i$. 


In this paper, we introduce and evaluate three types of tests: fixed-length, sequential, and two-phase. We show that under mild conditions, the maximum misclassification probability decreases exponentially for all tests. Specifically, the sequential test exhibits the highest exponential decay rates, whereas the fixed-length test features the simplest design. The two-phase test, which combines two fixed-length tests with different sample sizes, achieves an optimal balance between design complexity and performance.

\subsection{General Case}
We next consider a more general case with the null hypothesis, where the testing sequence is allowed to be generated i.i.d. from an arbitrary distribution $Q\in\calP(\calX)$ that is not close to any of the generating distribution of training sequences. Let $P_0\in\calP(\calX)$ be a distribution such that $P_0\notin\bigcup_{i\in[M]}\calS_{2\delta}(P_i)$. Under the null hypothesis, we have $Q\in\calS_{\delta}(P_0)$.

In this case, the definitions of $D_1(\delta,M)$ and $D_2(\delta,M)$ in \eqref{D1} and \eqref{D2} are updated as $\Bar{D}_1(\delta,M)$ and $\Bar{D}_2(\delta,M)$ as follows:
\begin{align}
&\Bar{D}_1(\delta,M)=\min_{i\in[M]}\min_{Q\in\bigcup_{j\in[0:M]\setminus\{i\}}\calS_{\delta}(P_j)} \mathrm{MMD}^2(Q,P_i),\label{D1_un}\\
&\Bar{D}_2(\delta,M)=\max_{i\in[0:M]}\max_{Q\in\calS_{\delta}(P_i)} \mathrm{MMD}^2(Q,P_i),\label{D2_un}
\end{align}
where $\Bar{D}_1(\delta,M)$ is the minimal inter-cluster distance between different set of distributions in $\{P_0,\bP\}$, and $\Bar{D}_2(\delta,M)$ is the maximal intra-cluster distance of a distributions set in $\{P_0,\bP\}$.
Similarly, we should have $\Bar{D}_1(\delta,M)>\Bar{D}_2(\delta,M)$ since otherwise no test can make reliable decision.

Analogous to the simple case, the task is to design a test $\Phi=(\tau, \phi_\tau)$, which includes a random stopping time $\tau\in\bbN$ and a mapping rule $\phi_\tau$ to decide among the following $M+1$ hypotheses:
\begin{itemize}
\item $\rmH_i$,~$i\in [M]$: the testing sequence is generated i.i.d. from the distribution that belongs to $\calS_\delta(P_i)$, i.e., $Q\in \calS_\delta(P_i)$.
\item $\rmH_\rmr$: the testing sequence is generated i.i.d. from a distribution in $\calS_\delta(P_0)$,  i.e., $Q\in\calS_\delta(P_0)$.
\end{itemize}
The null hypothesis $\rmH_\rmr$ models the case where the testing sequence is generated from a distribution that is vastly different from the generating distribution of any training sequence. This setting is reasonable since in practical classification tasks, a testing sample can be never seen before and does not belong to any of the training set.

To evaluate the performance of a test in this setting, under each non-null hypothesis $\rmH_i$ with $i\in[M]$, we consider the misclassification error probability in \eqref{Error_define}. Note that in the general case, $\beta_i(\phi_\tau|\bP,\delta)$ bounds the probability of the error event where the test incorrectly claims a wrong match $\rmH_j$ with $j\neq i$, or the test incorrectly favors the null hypothesis $\rmH_\rmr$. Furthermore, we also need the following false alarm probability under the null hypothesis:
\begin{align}
\nn&\rmP_{\rm{FA}}(\phi_\tau|\bP,\delta)\\*
&:= \max_{\substack{P_0\in\calP(\calX):\\P_0\notin\cup_{i\in[M]}\calS_{2\delta}(P_i)}}\max_{Q\in\calS_{\delta}(P_0)}
\bbP_\rmr\{\phi_\tau(X^\tau,\bY^{\alpha\tau})\neq \rmH_\rmr\},\label{FA_s1}
\end{align}
where under distribution $\bbP_\rmr$, the testing sequence is generated i.i.d. from an unknown distribution $Q\in\calS_{\delta}(P_0)$, where $P_0$ is not close to any of the distributions in $\bP$. Note that $\rmP_{\mathrm{FA}}(\phi_\tau|\bP,\delta)$ bounds the probability of the false alarm event where the test incorrectly claims that the testing sequence is generated from a distribution that is close to the generating distribution of a training sequence.

\subsection{MMD Metric}
Consistent with \cite{MMD}, we utilize the MMD metric proposed by \cite{gretton2012kernel} to develop our tests.
The MMD distance  between any two distributions $f_1$ and $f_2$ is defined as
\begin{align}
\nn\mathrm{MMD}^2(f_1,f_2)
&:=\bbE_{f_1f_1}[k(X,X')]-2\bbE_{f_1f_2}[k(X,Y)]\\
&+\bbE_{f_2f_2}[k(Y,Y')],
\end{align}
where $k(\cdot,\cdot)$ is a kernel function associated with the Reproducing Kernel Hilbert Spaces~\cite{sun2023kernel} and $(X,X',Y,Y')\sim f_1f_1f_2f_2$. Note that $\mathrm{MMD}^2(f_1,f_2)=0$ if and only if $f_1=f_2$. A usually adopted kernel function is the following Gaussian kernel function
\begin{align}
k(x,y):=\exp\left\{-\frac{(x-y)^2}{2\sigma_0^2}\right\}\label{Gaussiankernel},
\end{align}
where $(x,y)\in\calR^2$ and $\sigma_0\in\bbR_+$ is an positive real number. In this paper, we adopt the Gaussian kernel function in numerical examples to illustrate our theoretical results.

Given two sequences $x^{n_1}=[x_1,x_2,\ldots,x_{n_1}]$ and $y^{n_2}=[y_1,y_2,\ldots,y_{n_2}]$ sampled i.i.d. from $f_1$ and $f_2$, respectively, the MMD between the two sequences is defined as
\begin{align}
\nn\mathrm{MMD}^2(x^{n_1},y^{n_2})
\nn&:=\frac{1}{n_1(n_1-1)}\sum_{i,j\in[n_1],i\neq j}k(x_i,x_j)\\*
\nn&\qquad+\frac{1}{n_2(n_2-1)}\sum_{i,j\in[n_2],i\neq j}k(y_i,y_j)\\*
&\qquad-\frac{2}{n_1n_2}\sum_{i\in[n_1],j\in[n_2]}k(x_i,y_j)\label{MMDcompute}.
\end{align}
It was shown in~\cite[Lemma 6]{gretton2012kernel} that $\mathrm{MMD}^2(x^{n_1},y^{n_2})$ is an unbiased estimator of $\mathrm{MMD}^2(f_1, f_2)$.

\section{Main results for the simple case}
\label{Main}
We first recall and slightly generalize the results for the fixed-length test as given in \cite{li2018nonparametric}. Subsequently, we present our novel results for sequential and two-phase tests by showing that our sequential test has strictly better performance than the fixed-length test and our two-phase test strikes a good balance between performance and design complexity.

\subsection{Fixed-length Test}
\label{S-FLMT}

\subsubsection{Test Design and Asymptotic Intuition}


In a fixed-length test, the stopping time $\tau$ is set to $n\in\bbN$. Thus, the testing sequence has length $n$ and each training sequence has length $N=\alpha n$. For training sequences $\by^N=\{y_1^N,\ldots,y_M^N\}$ and a testing sequence $x^n=\{x_1,\ldots,x_n\}$, we define the following quantity:
\begin{align}
& i^*(x^n,\by^N):=\argmin_{i\in [M]}\mathrm{MMD}^2(x^n,y_i^N)\label{MinG}.
\end{align}
Note that $i^*(x^n,\by^N)$ denotes the index of training sequence that has the smallest MMD value with respect to the test sequence $x^n$.

Fix any positive real number $\lambda\in\bbR_+$. Our fixed-length test operates as follows:
\begin{equation}\label{FLTest}
\phi_n(x^n,\by^N)=
\rmH_i: \text{ if }i^*(x^n,\by^N)=i.
\end{equation}
In other words, our test $\phi_n$ claims that hypothesis $\rmH_i$ is true if $\mathrm{MMD}^2(x^n,y_i^N)$ is smallest.

We now provide an asymptotic explanation for the effectiveness of the test in \eqref{FLTest}. Fix any $i\in[M]$, assume that hypothesis $\rmH_i$ is true, i.e., the testing sequence $x^n$ is generated from a distribution $Q\in\calS_\delta(P_i)$. Asymptotically as the sample size $n\to\infty$, we have i) $\mathrm{MMD}^2(x^n,y_i^N)\to \mathrm{MMD}^2(Q,P_i)\leq D_2(\delta,M)$ and ii) for any $j\in\calM_i$, the MMD value $\mathrm{MMD}^2(x^n,y_j^N)\to \mathrm{MMD}^2(Q,P_j)\geq D_1(\delta,M)$. Using the condition that $D_1(\delta,M)>D_2(\delta,M)$, we conclude that the test can make a correct decision asymptotically.

\subsubsection{Theoretical Results and Discussions}

Consider a kernel function $k(x,y)$ with a finite maximum value, denoted as $K_0:=\max_{(x,y)\in\calX\times\calY}k(x,y)<\infty$. An example of such a kernel function is the Gaussian kernel described in \eqref{Gaussiankernel}, which meets this criterion with $K_0=1$.

For ease of notation, given any real number $D_2(\delta,M)<a<D_1(\delta,M)$, define the following exponent functions:
\begin{align}  
g_1(D_1,D_2)&:=\frac{(D_1(\delta,M)-D_2(\delta,M))^2}{ 32K_0^2\left(1+\frac{2}{\alpha}\right)}\label{g1},\\
g_2(a)&:=\frac{(D_1(\delta,M)-a)^2}{ 32K_0^2\left(1+\frac{1}{\alpha}\right)}\label{g2},\\
g_3(a)&:=\frac{(a-D_2(\delta,M))^2}{ 32K_0^2\left(1+\frac{1}{\alpha}\right)}\label{g3}.
\end{align}
As we shall show below, $g_1(D_1,D_2)$ is the misclassification exponent (the exponential decay rates of misclassification probability) for a fixed-length test, $g_1(D_1,D_2)$ and $g_2(\cdot)$ are used to characterize the misclassification exponent for a sequential test while $g_1(D_1,D_2)$ and $g_2(\cdot)$ are used to characterize the misclassification exponent for a two-phase test.

The following theorem characterizes the mismatch exponent for the fixed-length test in \eqref{FLTest}.
\begin{theorem}\label{FLMT}
Under any tuple of unknown distributions $\bP=\{P_1,P_2,\ldots,P_M\}$, the fixed-length test in \eqref{FLTest} ensures that for each $i\in[M]$, the misclassification exponent satisfies
\begin{align}
\liminf_{n\to\infty}-\frac{1}{n}\log \beta_i(\phi_n|\bP,\delta)
&\geq g_1(D_1,D_2).
\end{align}
\end{theorem}
For the case of $\alpha=1$, Theorem \ref{FLMT} was presented in \cite[Theorem 1]{li2018nonparametric}. For completeness, we provide a proof sketch in Appendix \ref{proof_of_FLMT}.

Theorem \ref{FLMT} implies that the misclassification probability decays exponentially with respect to the sample size $n$, where the exponential decay rate is a function of $D_1(\delta,M)-D_2(\delta,M)$. When $\delta$ increases, it follows from \eqref{D1} and \eqref{D2} that $D_1(\delta,M)$ decreases while $D_2(\delta,M)$ increases. Thus, increasing the uncertainty level of the generating distribution of the testing sequence leads to degraded performance with a smaller misclassification exponent.

We next discuss the effect of $\alpha$, the ratio between the lengths of each training and the testing sequence, on the misclassification exponent. Theorem \ref{FLMT} implies that the misclassification exponent increases as $\alpha$ increases. Thus, the test performs better if more training samples are available, which is consistent with our intuition. If the length of each training sequence is unlimited ($\alpha\to\infty$), the error exponent reaches its upper bound,  $\frac{(D_1(\delta,M)-D_2(\delta,M))^2}{ 32K_0^2}$, which corresponds to the case where the generating distribution of each training sequence is known.

In the next subsections, we generalize Theorem \ref{FLMT} to the case of sequential and two-phase tests, which have better performance with larger misclassification exponents.

\subsection{Sequential test}
\label{S-ST}

\subsubsection{Test Design and Asymptotic Intuition}
We then present a sequential test with a stopping time $\tau$ and a decision rule $\phi_\tau$. Let $N_0\in\bbN$ be a fixed integer. 
For the testing sequence $x^n$ and training sequences $\by^N=\{y_1^N,\ldots,y_M^N\}$, define the second minimal MMD metric as
\begin{align}
&h(x^n,\by^N):=\min_{i\in [M]:~i\neq i^*(x^n,\by^N)}\mathrm{MMD}^2(x^n,y_i^N)\label{secMinG}.
\end{align}

Recall that the ratio between the lengths of each training sequence and the testing sequence is assumed to be a constant $\alpha$. Given a positive real number $\lambda\in\bbR_+$, the stopping time $\tau$ of our sequential test is defined as follows:
\begin{align}
\tau&=\inf\{n\in\bbN:~n\geq N_0-1, h(x^n,\by^N)>\lambda\},\label{Taulength}
\end{align}
where $N_0$ is a design parameter of the sequential test to avoid stopping too early. At the stopping time $\tau$, we run the fixed-length test $\phi_n(x^n,\by^N)$ in \eqref{FLTest} with $(n,N)$ replaced by $(\tau,\alpha\tau)$. In a nutshell, with a sequential test, one continues collecting samples until a reliable decision can be made.

We next explain why the sequential test works from an asymptotic perspective. Fix any $i\in[M]$. Assume that hypothesis $\rmH_i$ is true.
As the sample size $n\to\infty$ and thus $N=\alpha n\to\infty$, it follows that i) $\mathrm{MMD}^2(x^n,y_i^{N})\to\mathrm{MMD}^2(Q,P_i)\leq D_2(\delta,M)$ and ii) for each $j\in\calM_i$, $\mathrm{MMD}^2(x^n,y_j^{N})\to\mathrm{MMD}^2(Q,P_j)\geq D_1(\delta,M)$. Thus, if $N_0$ is sufficiently large, when $D_2(\delta,M)<\lambda<D_1(\delta,M)$, our sequential test could always make a correct decision.

\subsubsection{Theoretical Results and Discussions}
\begin{theorem}\label{ST}
Under any tuple of unknown distributions $\bP=\{P_1,P_2,\ldots,P_M\}$, for any positive real number $\lambda\in\bbR_+$, our sequential test ensures that
\begin{enumerate}
\item  when $N_0$ is sufficiently large, the average stopping time satisfies
\begin{align}
\max_{i\in[M]}\bbE_{\bbP_i}[\tau]
&\le
\left\{
\begin{array}{ll}
N_0&\mathrm{if~}\lambda<D_1(\delta,M),\\
\infty&\mathrm{otherwise.}
\end{array}
\right.
\end{align}
\item for each $i\in[M]$, when $D_2(\delta,M)<\lambda<D_1(\delta,M)$, the misclassification exponent satisfies
\begin{align}
\nn&\liminf_{N_0\to\infty}-\frac{1}{\bbE_{\bbP_i}[\tau]}\log \beta_i(\phi_\tau|\bP,\delta)\\
&\geq \max\{g_1(D_1,D_2),g_3(\lambda)\}.
\end{align}
\item if $\lambda\leq D_2(\delta,M)$, the misclassification exponent  is the same as that of fixed-length test in Theorem \ref{FLMT}.
\end{enumerate}
\end{theorem}
The proof of Theorem \ref{ST} is provided in Appendix \ref{proof_of_ST}. To prove Theorem \ref{ST}, we first upper bound the average stopping time under each hypothesis. Subsequently, we use the similar idea, which was used to prove Theorem \ref{FLMT}, to bound the misclassification exponent of our sequential test, including calculating the expected value of the MMD metric and applying the McDiarmid's inequality~\cite{mcdiarmid1989method}.

It follows from Theorems \ref{FLMT} and \ref{ST} that the sequential test can achieve a larger misclassification exponent if the threshold $\lambda$ is chosen appropriately. Specifically, for any positive real number $0<\varepsilon<D_1(\delta,M)$, when $\lambda=D_1(\delta,M)-\varepsilon$, the misclassification exponent of our sequential test is at least $\max\Big\{\frac{(D_1(\delta,M)-D_2(\delta,M))^2}{ 32K_0^2\left(1+\frac{2}{\alpha}\right)},\frac{(D_1(\delta,M)-D_2(\delta,M)-\varepsilon)^2}{ 32K_0^2\left(1+\frac{1}{\alpha}\right)}\Big\}$ for any $\varepsilon\in(0,\frac{1}{2}(D_1(\delta,M)-D_2(\delta,M))$, which is strictly larger than the misclassification exponent of the fixed-length test in Theorem \ref{FLMT} when $\varepsilon\to 0$. The superior performance of our sequential test results from the use of $\lambda$ that determines the stopping time. In particular, $\lambda$ helps to reduce the ambiguity among hypotheses by allowing one to collect as many samples as enough to make a reliable decision. However, if $0<\lambda\leq D_2(\delta,M)$, our sequential test always stops at $\tau=N_0$, reduces to a fixed-length test and thus achieves the same misclassification exponent as in Theorem \ref{FLMT}.

Although our sequential test has superior performance by having larger misclassification exponent, one significant cost is the high computational complexity. As shown in Fig. \ref{complexity_detection_error_known}, the sequential test takes much longer than the fixed-length test to make a decision. This is because one needs to check the stopping criterion after collecting each data sample. One might wonder whether we can achieve the superior performance of the sequential test with design complexity similar to a fixed-length test. In the next subsection, we answer this question affirmatively by proposing a test with two possible stopping times and prove that it strikes a good balance between design complexity and test performance.

\subsection{Two-phase test}
\label{S-AFLMT}
\subsubsection{Test Design and Asymptotic Intuition}
Fix two integers $(K,n)\in\bbN^2$ and two positive real numbers $(\lambda_1,\lambda_2)\in\bbR_+^2$. 
Our two-phase test has a random stopping time $\tau$ which can only take two possible values, either $n$ or $Kn$. For any training sequences $\by^{KN}=\{y_1^{KN},\ldots,y_M^{KN}\}$ and the testing sequence $x^{Kn}$, the stopping time $\tau$ satisfies
\begin{align}\label{Taulength2}
\tau:=\left\{
\begin{aligned}
n & \text{ if }h(x^n,\by^N)>\lambda ;\\
Kn & \text{ otherwise},
\end{aligned}
\right.
\end{align}
where $x^n$ is the first $n$ samples of $x^{Kn}$ and $\by^N$ are the first $N$ samples of $\by^{KN}$. At the stopping time $\tau$, our two-phase test operates as follows. When $\tau=n$, we apply the fixed-length test $\phi_n(x^n,\by^N)$ in \eqref{FLTest}; when $\tau=Kn$, we apply the fixed-length test $\phi_n(x^n,\by^N)$ in \eqref{FLTest} with $n$ replaced by $Kn$.


To summarize, our two-phase testing procedure is as follows. In the first phase, the test gathers $n$ samples and conducts a fixed-length test with an option to reject. If the criterion $h(x^n, \by^N) < \lambda$ is met, indicating that the hypothesis cannot be reliably determined, the test calls for additional samples. Upon making a rejection decision in the first phase, the second phase begins, where an additional $(K-1)n$ samples are collected. A final fixed-length test, which does not include a rejection option, is then performed to reach a conclusive decision. Note that $\lambda$ functions similarly in the two-phase test as it does in the sequential test. The rationale behind the effectiveness of the two-phase test mirrors that of the sequential test and is thus omitted.

\subsubsection{Theoretical Results and Discussions}
\begin{theorem}\label{AFLMT}
Under any tuple of unknown generating distributions $\bP=\{P_1,P_2,\ldots,P_M\}$, for any positive real numbers $\lambda\in\bbR_+$, our two-phase test ensures that
\begin{enumerate}
\item  when $n$ is sufficiently large, the average stopping time satisfies
\begin{align}
\max_{i\in[M]}\bbE_{\bbP_i}[\tau]&\leq
\left\{
\begin{array}{ll}
n+1 &\mathrm{if~}\lambda<D_1(\delta,M),\\
Kn&\mathrm{otherwise.}
\end{array}
\right.
\end{align}
\item if $D_2(\delta,M)<\lambda<D_1(\delta,M)$, for each $i\in[M]$, the misclassification exponent satisfies
\begin{align}
\nn&\liminf_{n\to\infty}-\frac{1}{\bbE_{\bbP_i}[\tau]}\log \beta^{\mathrm{tp}}_i(\phi_{\tau}|\bP,\delta)\\
\nn&\geq \min\big\{ \max\{g_1(D_1,D_2),g_3(\lambda)\}, Kg_1(D_1,D_2)\big\}
\end{align}
\item if $\lambda\geq D_1(\delta,M)$ or $0<\lambda< D_2(\delta,M)$, the misclassification exponent of our two-phase test reduces to that of fixed-length test in Theorem \ref{FLMT}.
\end{enumerate}
\end{theorem}

It follows from Theorem \ref{AFLMT} that the misclassification exponent of our two-phase test takes a minimization between two exponent functions. The first item inside the minimization is the error exponent during the first phase, which is the same as that given in Theorem \ref{ST} because the test of the first phase has the same decision rule as the sequential test. The second item inside the minimization comes from the fixed-length test during the second phase with using $Kn$ samples from the test sequence and $KN$ samples from each training sequence, which is the same as Theorem \ref{FLMT} except a multiplicative factor of $K$. The proof of Theorem \ref{AFLMT} is thus omitted since it combines the proofs of Theorems \ref{FLMT} and \ref{ST} as explained above.


By comparing Theorem \ref{AFLMT} with Theorems \ref{FLMT} and \ref{ST}, we find that the two-phase test bridges the gap between fixed-length and sequential tests. Theorem \ref{AFLMT} shows that the two-phase test simplifies to a fixed-length test when $K=1$. Conversely, with a sufficiently large $K$, the error exponent of the two-phase test matches that of the sequential test. Numerical examples in Fig. \ref{error_complexity_known} demonstrate that the two-phase test balances performance and complexity effectively compared to fixed-length and sequential tests.

\section{Main results for the general case}
\label{Main_un}

In this section, we consider the case where the null hypothesis can be true such that the generating distribution of the testing sequence may be vastly different from the generating distribution of any training sequence.

\subsection{Fixed-length Test}
\label{S-FLMT-un}

\subsubsection{Test Design and Asymptotic Intuition}

Using the definitions of $i^*(x^n,\by^N)$ in \eqref{MinG}, for any positive real number $\lambda\in\bbR_+$, our fixed-length test operates as follows:
\begin{align}\label{FLTest_un}
\nn&\phi_n(x^n,\by^N)\\*
&=
\left\{
\begin{array}{ll}
\rmH_i &\text{if}~i^*(x^n,\by^N)=i \mathrm{~and~}  \mathrm{MMD}^2(x^n,y_i^N)<\lambda;\\
\rmH_\mathrm{r}&\text{otherwise}.
\end{array}
\right.
\end{align}
Specifically, the test $\phi_n$ claims that hypothesis $\rmH_i$ is true if $\mathrm{MMD}^2(x^n,y_i^N)$ is smallest and is also smaller than the threshold $\lambda$, while the test outputs the null hypothesis $\rmH_\rmr$ if the minimal MMD scoring function $\min_{j\in[M]}\mathrm{MMD}^2(x^n,y_j^N)$ is larger than the threshold $\lambda$.

We next explain the asymptotic intuition why the test in \eqref{FLTest_un} works. For each non-null hypothesis, the intuition is very similar as the that in the simple case, which states that if the threshold satisfies $\Bar{D}_2(\delta,M)<\lambda<\Bar{D}_1(\delta,M)$, the test can make correct decision. For the null hypothesis $\rmH_\rmr$, as the sample size $n\to\infty$, it follows that for each $i\in[M]$, $\mathrm{MMD}^2(x^n,y_i^N)\to\mathrm{MMD}^2(Q,P_i)\geq \Bar{D}_1(\delta,M)$. Thus, as long as $\lambda<\Bar{D}_1(\delta,M)$, the correct decision $\rmH_\rmr$ would be made asymptotically. In the next subsection, we explicitly characterize the exponents of all three error probabilities.

\subsubsection{Theoretical Results and Discussions}
Analogously to the simple case, for ease of notation, given any real number $\Bar{D}_2(\delta,M)<a<\Bar{D}_1(\delta,M)$, we define $\Bar{g}_1(\Bar{D}_1,\Bar{D}_2)$, $\Bar{g}_2(a)$ and $\Bar{g}_3(a)$ with the same expressions as those in \eqref{g1}-\eqref{g3} by replacing $(D_1(\cdot),D_2(\cdot))$ with $(\Bar{D}_1(\cdot),\Bar{D}_2(\cdot))$. Define the following set
\begin{align}
\calT:=\{a\in\bbR_+:~\Bar{D}_2(\delta,M)<a<\Bar{D}_1(\delta,M)\}
\label{def:calT}.
\end{align}

\begin{theorem}\label{FLMT_un}
Under any tuple of unknown distributions $\bP=\{P_1,P_2,\ldots,P_M\}$, for any positive real number $\lambda\in\bbR_+$, the test in \eqref{FLTest_un} ensures that
\begin{enumerate}
\item for each $i\in[M]$, when $\lambda\in\calT$, the misclassification exponent satisfies
\begin{align}
\nn&\liminf_{n\to\infty}-\frac{1}{n}\log \beta_i(\phi_n|\bP,\delta)\\*
&\geq\min\big\{\max\{\Bar{g}_1(\Bar{D}_1,\Bar{D}_2),\Bar{g}_2(\lambda)\},\Bar{g}_3(\lambda)\big\}.\label{FLm_un}
\end{align}
If $\lambda\geq \Bar{D}_1(\delta,M)$, the misclassification exponent is the same as that of fixed-length test in the simple case except that $(D_1(\cdot),D_2(\cdot))$ are replaced with $(\Bar{D}_1(\cdot),\Bar{D}_2(\cdot))$. If $\lambda\leq \Bar{D}_2(\delta,M)$, the misclassification exponent equals zero.
\item when $\lambda<\Bar{D}_1(\delta,M)$, the false alarm exponent satisfies
\begin{align}
\liminf_{n\to\infty}-\frac{1}{n}\log \rmP_{\mathrm{FA}}(\phi_n|\bP,\delta)
&\geq \Bar{g}_2(\lambda).
\end{align}
If $\lambda\geq \Bar{D}_1(\delta,M)$, the false alarm exponent equals zero.
\end{enumerate}
\end{theorem}
The proof of Theorem \ref{FLMT_un} is provided in Appendix \ref{proof_of_FLMT_Un}, which generalizes the proof of Theorem \ref{FLMT} by considering whether the null hypothesis is true.

It follows from Theorem \ref{FLMT_un} that the misclassification exponent of the fixed-length test is a minimization of two terms. This is because under each each non-null hypothesis, a misclassification event occurs if the testing sequence is incorrectly classified or if the test makes a false reject error by incorrectly claiming the null hypothesis. Fix any $i\in[M]$ and thus hypothesis $\rmH_i$. The event of incorrect classification occurs when the minimal MMD metric is corresponding to the testing sequence $Y_j^N$ with $j\in\calM_i$ and is also smaller than the threshold $\lambda$, which leads to the first term that involves a maximization of two exponent functions $(\barg_1(\barD_1,\barD_2),\barg_2(\lambda))$. The false reject event occurs if the minimal MMD metric is greater than the threshold $\lambda$, which leads to the exponent function $\barg_3(\lambda)$. 

Theorem \ref{FLMT_un} implies that 
the threshold $\lambda$ tradeoffs the two items inside misclassification exponent, and also tradeoffs the misclassification and false alarm exponents.  Specifically, if $\lambda$ tends to $\Bar{D}_1(\delta,M)$, $\Bar{g}_2(\lambda)$ tends to zero and $\Bar{g}_3(\lambda)$ tends to $\frac{(\Bar{D}_1(\delta,M)-\Bar{D}_2(\delta,M))^2}{ 32K_0^2\left(1+\frac{1}{\alpha}\right)}$. In this case, the misclassification exponent is maximized, while the false alarm exponent tends to zero. On the other hand, if $\lambda$ tends to $\Bar{D}_2(\delta,M)$, according to the definitions of $\barg_2(\lambda)$ and $\barg_3(\lambda)$, the false alarm exponent is maximized, while the misclassification exponent tends to zero. As we shall below, the sequential test resolves the above exponent tradeoff and achieves better test performance.

Comparing Theorem \ref{FLMT_un} and Theorem \ref{FLMT}, we conclude that the misclassification exponent of the fixed-length test in the general case is no larger than that in the simple case. Specifically, if $\barg_2(\lambda)\leq \barg_1(\Bar{D}_1,\Bar{D}_2)$, the misclassification exponent in \eqref{FLm_un} equals to $\min\big\{\Bar{g}_1(\Bar{D}_1\Bar{D_1}),\Bar{g}_3(\lambda)\big\}$, which is no larger than $\Bar{g}_1(\Bar{D}_1,\Bar{D}_2)$. Otherwise, if $\barg_2(\lambda)> \barg_1(\Bar{D}_1,\Bar{D}_2)$, \eqref{FLm_un} equals to $\min\big\{\Bar{g}_2(\lambda),\Bar{g}_3(\lambda)\}$, which is no larger than $\frac{(\Bar{D}_1(\delta,M)-\Bar{D}_2(\delta,M))^2}{ 64K_0^2\left(1+\frac{1}{\alpha}\right)}$. Combing the results above and using the fact that $\Bar{D}_1(\delta,M)\leq D_1(\delta,M)$ and $\Bar{D}_2(\delta,M)\geq D_2(\delta,M)$ (cf. \eqref{D1}-\eqref{D2}, \eqref{D1_un}-\eqref{D2_un}), we theoretically verify that the misclassification exponent of the general case can be smaller.  As shown in Fig. \ref{detection_error_delta}, the misclassification exponent of the fixed-length test in the simple case can be strictly larger. Thus, there is a penalty of not knowing whether the null hypothesis is true. Such a result is consistent with our intuition since in the general case, the additional null hypothesis makes the task more challenging.

\subsection{Sequential test}\label{S-ST-un}
\subsubsection{Test Design and Asymptotic Intuition}
We next present a sequential test with a stopping time $\tau$ and a decision rule $\phi_\tau$. Recall $N_0\in\bbN$ is a fixed integer. 
Given two positive real numbers $(\lambda_1,\lambda_2)\in\bbR_+^2$, 
the stopping time $\tau$ of our sequential test is defined as follows:
\begin{align}
\nn\tau&=\inf\{n\in\bbN:~n\geq N_0-1, \min_{j\in[M]}\mathrm{MMD}^2(x^n,y_j^{N})<\lambda_1\\
&\text{ and }h(x^n,\by^N)>\lambda_2,\text{ or }\min_{j\in[M]}\mathrm{MMD}^2(x^n,y_j^{N})>\lambda_2\}.\label{Taulength_un}
\end{align}
At the stopping time $\tau$, we run the fixed-length test $\phi_n(x^n,\by^N)$ in \eqref{FLTest_un} with $(n,N,\lambda)$ replaced by $(\tau,\alpha\tau,\lambda_1)$. We let $\lambda_1\leq \lambda_2$ to avoid the case where the test reduces to a fixed-length test with sample size $N_0-1$.

We next explain the asymptotic intuition why the sequential test works. Under each non-null hypothesis, the intuition is very similar as that in the simple case. Under the null hypothesis $\rmH_\rmr$, as the sample size $n\to\infty$, it follows that for each $i\in[M]$, $\mathrm{MMD}^2(x^n,y_i^{N})\to\mathrm{MMD}^2(Q,P_i)\geq \Bar{D}_1(\delta,M)$. Thus, if $N_0$ is sufficiently large and  $\lambda_2<\Bar{D}_1(\delta,M)$, the null hypothesis could be correctly decided asymptotically.

\subsubsection{Theoretical Results and Discussions}
Recall the definition of $\calT$ in \eqref{def:calT}.
\begin{theorem}\label{ST_un}
Under any tuple of unknown distributions $\bP=\{P_1,P_2,\ldots,P_M\}$, for any positive real numbers $(\lambda_1,\lambda_2)\in\bbR_+^2$ such that $\lambda_1\leq\lambda_2$, our sequential test ensures that
\begin{enumerate}[1)]
\item  when $N_0$ is sufficiently large, the average stopping time satisfies
\begin{align}
\max_{i\in[M]}\bbE_{\bbP_i}[\tau]
&\le
\left\{
\begin{array}{ll}
N_0&\mathrm{if~}(\lambda_1,\lambda_2)\in\calT^2,\\
\infty&\mathrm{otherwise.}
\end{array}
\right.\\
\bbE_{\bbP_\rmr}[\tau]
&\le
\left\{
\begin{array}{ll}
N_0&\mathrm{if~}\lambda_2<\Bar{D}_1(\delta,M),\\
\infty&\mathrm{otherwise.}
\end{array}
\right.
\end{align}
\item for each $i\in[M]$, if $(\lambda_1,\lambda_2)\in\calT^2$, the misclassification exponent satisfies
\begin{align}
\lim_{N_0\rightarrow\infty}-\frac{\log \beta_i^{\mathrm{seq}}(\phi_\tau|\bP,\delta)}{\bbE_{\bbP_i}[\tau]}
&\geq  \Bar{g}_3(\lambda_2).
\end{align}
\item for any $\lambda_1<\Bar{D}_1(\delta,M)$, the false alarm exponent satisfies
\begin{align}
\liminf_{N_0\to\infty}-\frac{1}{\bbE_{\bbP_\rmr}[\tau]}\log \rmP^{\rmseq}_{\rm{FA}}(\phi_\tau|\bP,\delta)
\geq \Bar{g}_2(\lambda_1).
\end{align}
\end{enumerate}
\end{theorem}

The proof of Theorem \ref{ST_un} is similar to the proof of Theorem \ref{ST} and provided in Appendix \ref{proof_of_ST_un} for completeness.

Comparing Theorems \ref{ST} and \ref{ST_un}, we conclude that the penalty of not knowing whether the null hypothesis is true also holds the sequential test. This result is intuitive because $\max\{g_1(D_1,D_2),g_3(\lambda)\}\geq g_3(\lambda)\geq \barg_3(\lambda_2)$ if $\lambda=\lambda_2$ since $\Bar{D}_1(\delta,M)\leq D_1(\delta,M)$ and $\Bar{D}_2(\delta,M)\geq D_2(\delta,M)$. The numerical results in Fig. \ref{detection_error_known} and Fig. \ref{detection_error_unknown}-(a) verify this conclusion. 

Comparing Theorems \ref{FLMT_un} and \ref{ST_un}, we conclude that the sequential test achieves larger misclassification and false alarm exponents and the sequential test resolves the tradeoff between the misclassification exponent and the false alarm exponent of the fixed-length test. Specifically, the misclassification and false alarm error exponents of the sequential test can be maximized simultaneously by setting $\lambda_1=\Bar{D}_2(\delta,M)+\varepsilon$ and $\lambda_2=\Bar{D}_1(\delta,M)-\varepsilon$ for any $\varepsilon\in\big(0,(\Bar{D}_1(\delta,M)-\Bar{D}_2(\delta,M))/2\big)$. In this case, both the misclassification and false alarm exponent equal $\frac{(\Bar{D}_1(\delta,M)-\Bar{D}_2(\delta,M)-\varepsilon)^2}{ 32K_0^2\left(1+\frac{1}{\alpha}\right)}$. Analogous to the simple case, the superior performance of the sequential test results from the freedom to stop at any possible time and the uses of two different thresholds $(\lambda_1,\lambda_2)$. However, such a test design is complicated since one needs to check the stopping criterion after collecting each data sample. In the next subsection, we propose a test with two possible stopping times and prove that the test strikes a good tradeoff between design complexity and test performance.

\subsection{Two-phase test}\label{S-AFLMT_un}
\subsubsection{Test Design and Asymptotic Intuition}
Fix two integers $(K,n)\in\bbN^2$. Our two-phase test has a random stopping time $\tau$ which only takes two possible values: $n$ or $Kn$. Given three positive real numbers $(\lambda_1,\lambda_2,\lambda_3)\in\bbR_+^3$, the stopping time of our two-phase test satisfies
\begin{equation}\label{Taulength_un2}
\tau:=\left\{
\begin{aligned}
n :& \text{ if }\min_{j\in[M]}\mathrm{MMD}^2(x^n,y_j^{N})<\lambda_1\\
&\qquad\qquad\text{ and }h(x^n,\by^N)>\lambda_2,\\
&\quad\text{ or }\min_{j\in[M]}\mathrm{MMD}^2(x^n,y_j^{N})>\lambda_2,\\
Kn :& \text{ otherwise}.
\end{aligned}
\right.
\end{equation}
Specifically, the test stops at $\tau=n$ if either a non-null hypothesis or the null hypothesis is believed to be true. Otherwise, the test proceeds to collect $(K-1)n$ additional samples from the testing sequence and $\alpha(K-1)n$ additional samples from each training sequence.

At the stopping time $\tau$, given training sequences $\by^{\alpha\tau}$ and the testing sequence $x^{\tau}$, our two-phase test operates as follows. When $\tau=n$, we use the fixed-length test $\phi_n(x^n,\by^{N})$ in \eqref{FLTest_un} with $\lambda$ replaced by $\lambda_1$. When $\tau=Kn$, we use the fixed-length test $\phi_n(x^n,\by^{N})$ in \eqref{FLTest_un} with $(n,\lambda)$ replaced by $(Kn,\lambda_3)$. 
In summary, our two-phase test involves two fixed-length tests using $n$ samples initially and $Kn$ samples subsequently. Therefore, the reasoning behind the effectiveness of the two-phase test is analogous to that of the fixed-length test and is not further detailed here.


Note that when $\lambda_1>\lambda_2$, we always have $\tau=n$ and thus the two-phase test reduces to the fixed-length test. To avoid degenerate cases, we require that $\lambda_1\leq\lambda_2$.

\subsubsection{Theoretical Results and Discussions}
Recall the definition of $\calT$ in \eqref{def:calT}.
\begin{theorem}\label{E-AFLMT}
Under any tuple of unknown distributions $\bP=\{P_1,P_2,\ldots,P_M\}$, for any positive real numbers $(\lambda_1,\lambda_2,\lambda_3)\in\bbR_+^3$ such that $\lambda_1\leq\lambda_2$, our two-phase test ensures that
\begin{enumerate}
\item when $n$ is sufficiently large, the average stopping time satisfies
\begin{align}
\max_{i\in[M]}\bbE_{\bbP_i}[\tau]&\leq
\left\{
\begin{array}{ll}
n+1 &\mathrm{if~}(\lambda_1,\lambda_2)\in\calT^2,\\
Kn &\mathrm{otherwise.}
\end{array}
\right.\\
\bbE_{\bbP_\rmr}[\tau]&\leq 
\left\{
\begin{array}{ll}
n+1 &\mathrm{if~}\lambda_2<\Bar{D}_1(\delta,M),\\
Kn &\mathrm{otherwise.}
\end{array}
\right.
\end{align}
\item when $(\lambda_1,\lambda_2)\in\calT^2$, for each $i\in[M]$, the misclassification exponent satisfies
\begin{align}
\nn&\liminf_{n\to\infty}-\frac{1}{\bbE_{\rmP_i}[\tau]}\log \beta^\rmtp_i(\phi_n|\bP,\delta)\geq\\
& \min\big\{\Bar{g}_3(\lambda_2), K\Bar{g}_3(\lambda_3),
\max\{
K\Bar{g}_1(\Bar{D}_1,\Bar{D}_2),K\Bar{g}_2(\lambda_3)
\}
\big\}
\end{align}
If $\lambda_1\geq \Bar{D}_1(\delta,M)$ or $\lambda_2\leq \Bar{D}_2(\delta,M)$, the misclassification exponent is the same as that of fixed-length test with $\lambda_3$ playing the role of $\lambda$.
\item when $\lambda_1<\Bar{D}_1(\delta,M)$, the false alarm exponent satisfies
\begin{align}
\nn&\liminf_{n\to\infty}-\frac{1}{\bbE_{\rmP_\rmr}[\tau]}\log \rmP_\mathrm{FA}^\rmtp(\phi_n|\bP,\delta)\\
&\geq \min\big\{\Bar{g}_2(\lambda_1),K\Bar{g}_2(\lambda_3)\big\}.
\end{align}
If $\lambda_1\geq \Bar{D}_1(\delta,M)$, the false alarm exponent is the same as that of fixed-length test with $\lambda_3$ playing the role of $\lambda$.
\end{enumerate}
\end{theorem}

The proof of Theorem \ref{E-AFLMT} extends the argument of Theorem \ref{AFLMT} to a more general scenario. As it closely follows the proofs of Theorems \ref{AFLMT}, \ref{FLMT_un}, and \ref{ST_un}, it is omitted here for brevity.

Analogous to the simple case, the performance of the two-phase test bridges over the performance of the fixed-length test in Theorem \ref{FLMT_un} and the sequential test in Theorem \ref{ST_un}.  Specifically, if we choose $K=1$ and set $\lambda_1=\lambda_2=\lambda_3=\lambda$, the exponents in Theorem \ref{E-AFLMT} reduce to that of the fixed-length test in Theorem \ref{FLMT_un}. On the other hand, if we set $K$ be large enough, the exponents of the two-phase test is the same as that of the sequential test in Theorem \ref{ST_un}.

Furthermore, it follows from Theorems \ref{E-AFLMT} and \ref{FLMT} that
the misclassification exponent of the two-phase test in the general case is no larger than that in the simple case. This is because $\min\{K\Bar{g}_3(\lambda_3),
\max\{
K\Bar{g}_1(\Bar{D}_1,\Bar{D}_2),K\Bar{g}_2(\lambda_3)
\}\}\leq K\Bar{g}_1(\Bar{D}_1,\Bar{D}_2)$ by following the similar analysis as the third remark of Theorem \ref{FLMT_un}. Thus, there is also a penalty of not knowing whether the null hypothesis is true for the two-phase test. In Fig. \ref{detection_error_delta}, we provide a numerical example to demonstrate the penalty on the misclassification exponent. The penalty can be eliminated if we set $\lambda_2=\lambda_3=\lambda=\Bar{D}_1(\delta,M)-\varepsilon$ while $\lambda_1=\Bar{D}_2(\delta,M)+\varepsilon$ for $\varepsilon\in(0,(\Bar{D}_1(\delta,M)-\Bar{D}_2(\delta,M))/2)$. However, if $\lambda_3\to \Bar{D}_1(\delta,M)$, the false alarm exponent tends to zero. Thus, to ensure both misclassification and false alarm probabilities decay exponentially fast, the penalty on the misclassification exponent always exists.

\section{Numerical Results}
\label{simulation}
In this section, we simulate the performance of our proposed tests. Consider a tuple of generating distributions $\bP=\{P_1,P_2,\ldots,P_M\}$ for the training sequence. Fix any $i\in[M]$. Under the non-null hypothesis $\rmH_i$, the testing sequence is generated from a distribution $Q$ that falls into the uncertainty set around $P_i$, i.e., $Q\in\calS_{\delta}(P_i)$. Under the null hypothesis, we have $Q\in\calS_{\delta}(P_0)$, where $P_0$ is not close to any set of distributions generated by $\bP$. 

All distributions $\bP$ and $P_0$ are set to be Gaussian distributions with different means and the same variance one. For each $i\in[M]$, $P_i=\calN(1.5(i-1),1)$ and $P_0=\calN(1.5M,1)$. For any $i\in[0:M]$, we use $\mu_i$ to denote the mean of distribution $P_i$ and set the distribution uncertainty set as $\calS_\delta(P_i):=\cup_{\epsilon\in[\mu_i-0.1,\mu_i+0.1]}\calN(\epsilon,1)$.   Under the above setting, we calculate the values of $D_1(\delta,M), D_2(\delta,M), \Bar{D}_1(\delta,M),\Bar{D}_2(\delta,M)$ the definitions in using \eqref{D1}, \eqref{D2}, \eqref{D1_un},\eqref{D2_un}. Finally, to calculate the MMD metric~\eqref{MMDcompute}, the Gaussian kernel in \eqref{Gaussiankernel} with $\sigma=1$ is used. Unless otherwise stated, we set $M = 10$ and $K=2$.

\begin{figure}[htbp]
\centering
\includegraphics[height=0.3\textwidth]{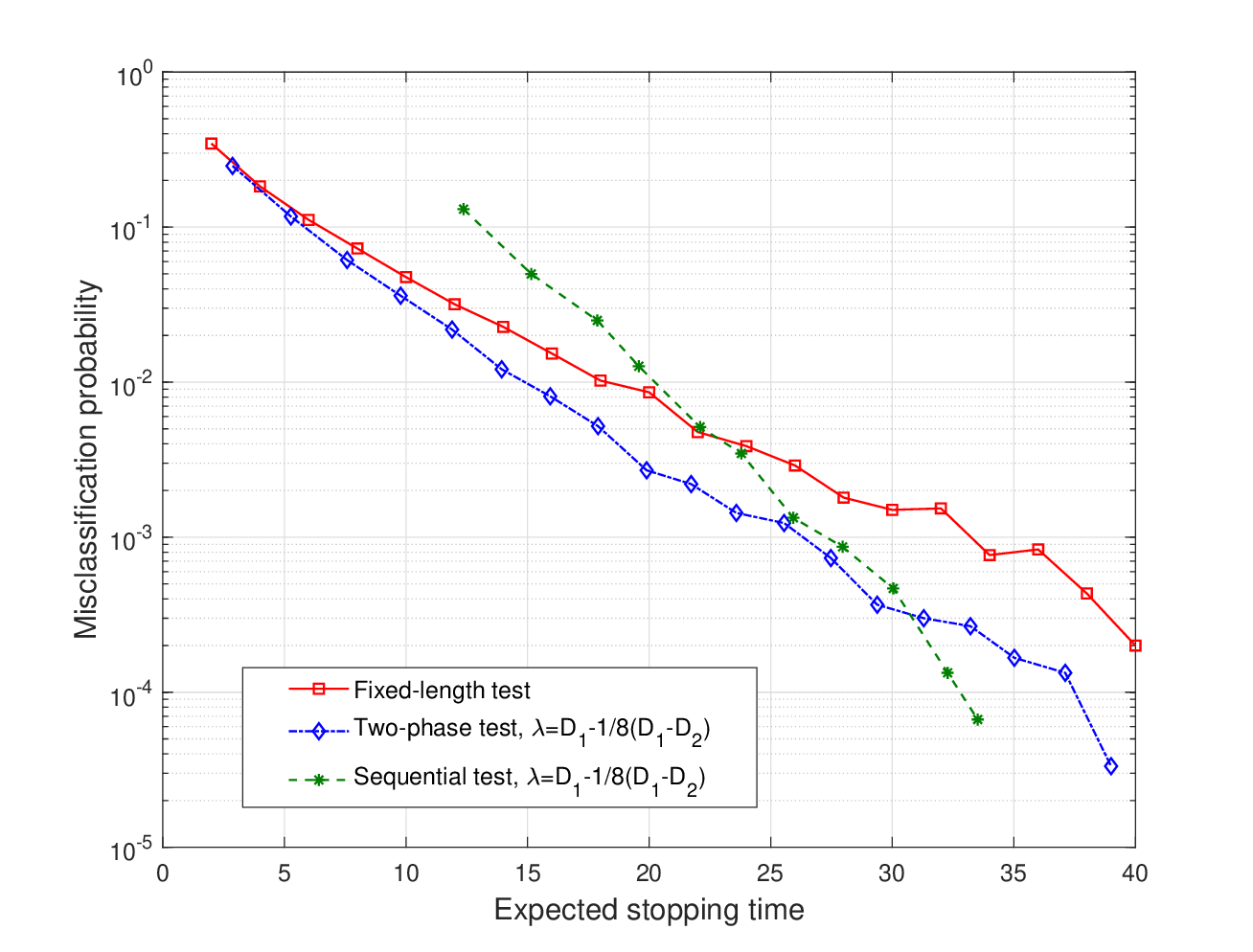}
\caption{Plot of simulated misclassification probabilities when $M=10$ and hypothesis $\rmH_1$ is true for our fixed-length test in Section \ref{S-FLMT}, our sequential test in Section \ref{S-ST}, and our two-phase test in Section \ref{S-AFLMT}. As observed, both our two-phase and sequential tests achieve better performance than the fixed-length test.}
\label{detection_error_known}
\end{figure}

In Fig. \ref{detection_error_known}, for the simple case, we plot the simulated misclassification probabilities of fixed-length, sequential, and two-phase tests in Section \ref{Main}. We set $n\in [40]$ as the sample length of the fixed-length test and the first phase of the two-phase test. For the sequential test, we set $N_0\in[40]$ where $N_0-1$ is the starting length. The expected stopping length of the fixed-length test is thus $n$ while those of sequential and two-phase tests are obtained by averaging the stopping times over $50000$ independent runs of our tests. As observed from Fig. \ref{detection_error_known}, our two-phase test and sequential test achieve better performance than the fixed-length test. As the expected stopping time increases, the sequential test achieves best performance while the two-phase test achieves relatively better performance at very small expected stopping times. The latter is because when $N_0$ is small, the sequential test tends to stop early and makes wrong decisions. The above numerical results are consistent with our theoretical analyses in Section \ref{Main}.

\begin{figure}[htbp]
\centering
\includegraphics[height=0.3\textwidth]{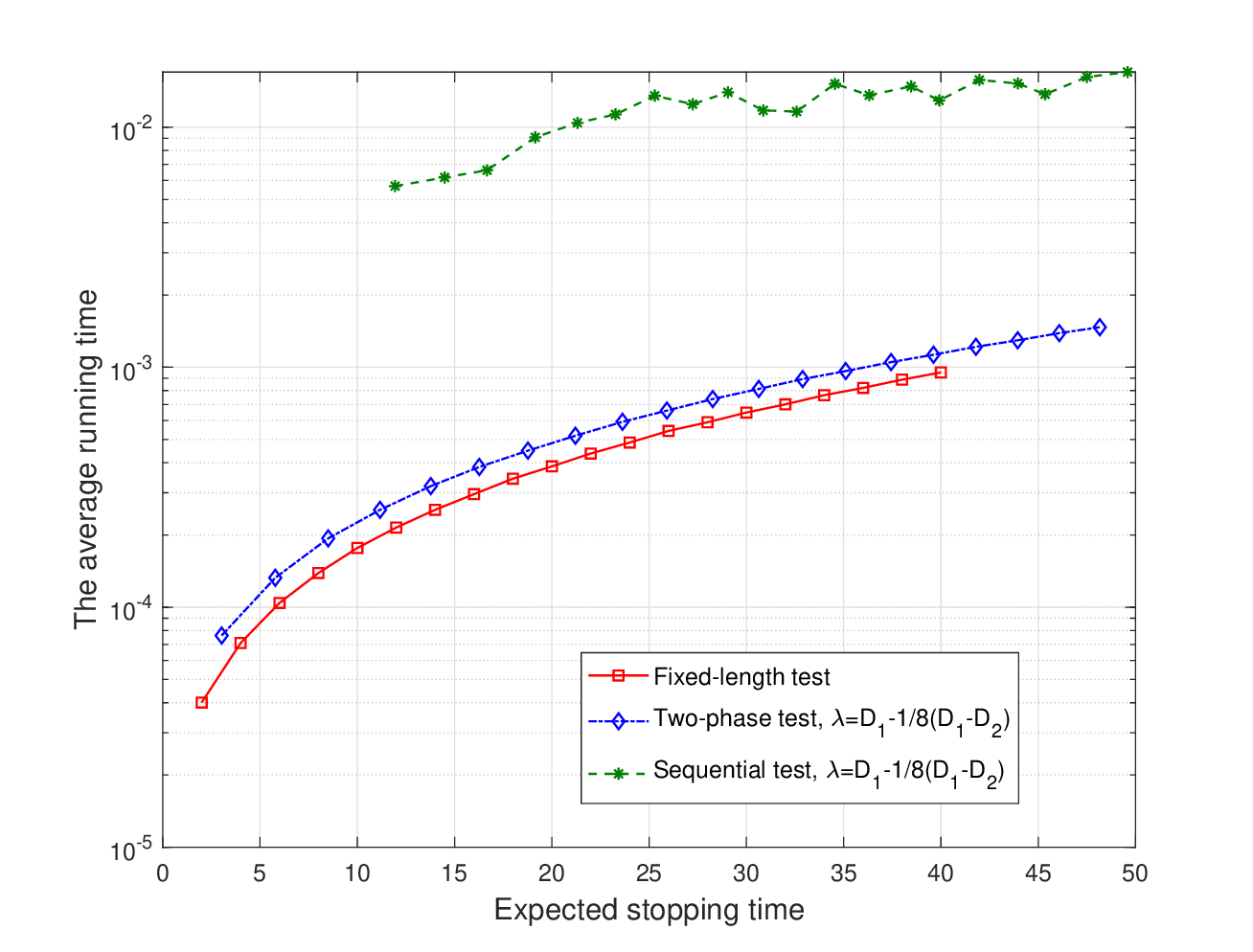}
\caption{Plot of simulated average running times of our fixed-length, sequential, and two-phase tests in Section \ref{Main} as a function of the expected stopping time for the same setting as Fig. \ref{detection_error_known}. As observed, our fixed-length and two-phase tests have much smaller running times than the sequential test.}
\label{complexity_detection_error_known}
\end{figure}

\begin{figure}[htbp]
\centering
\includegraphics[height=0.3\textwidth]{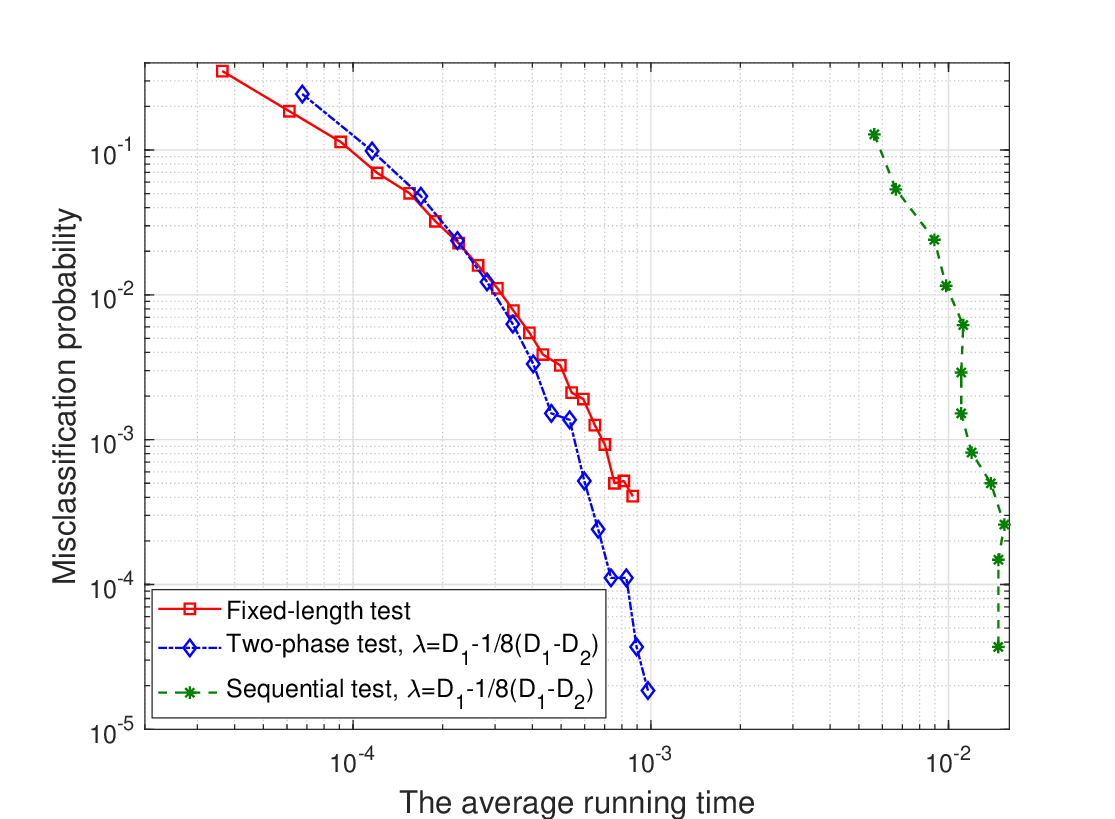}
\caption{Plot of simulated misclassification probabilities of our fixed-length, sequential, and two-phase tests in Section \ref{Main} as a function of the average running time for the same setting as Fig. \ref{detection_error_known}. }
\label{error_complexity_known}
\end{figure}

In Fig. \ref{complexity_detection_error_known}, for the same setting as Fig. \ref{detection_error_known}, we plot the average running times of all three tests as a function of the expected stopping time, while in Fig. \ref{error_complexity_known}, we plot simulated misclassification probabilities as a function of the simulated average running time. As observed from Fig. \ref{complexity_detection_error_known}, the sequential test is most computationally complicated while the two-phase test and the fixed-length test have roughly the same computational complexity. Furthermore, Fig. \ref{error_complexity_known} indicates that when the misclassification probability becomes smaller, the two-phase test performs better than the fixed-length test. The above results are consistent with our theoretical findings and confirm that our two-phase test strikes a good tradeoff between the performance and the design complexity. 

\begin{figure}[htbp]
\centering
\subfigure[Non-null hypothesis]{
\includegraphics[height=0.3\textwidth]{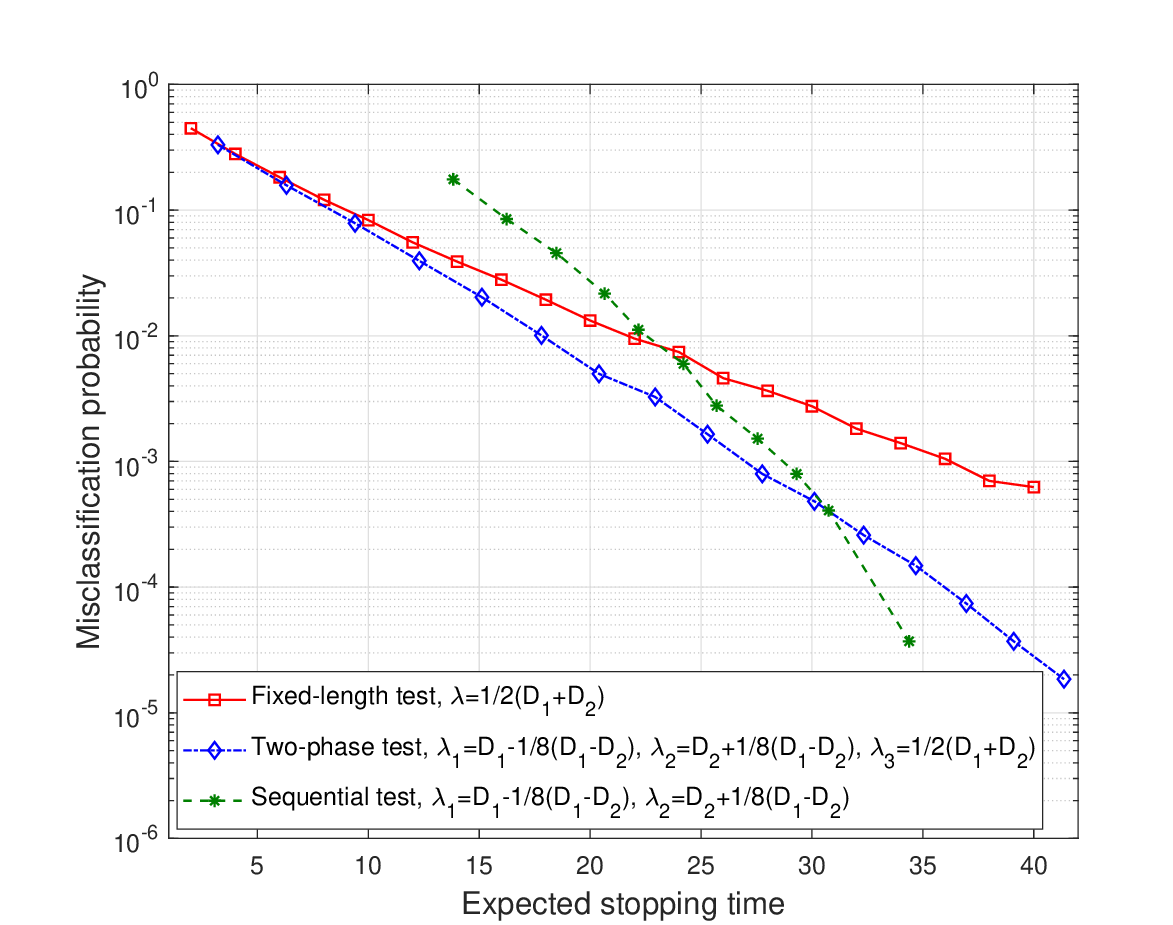}
}
\subfigure[Null hypothesis]{
\includegraphics[height=0.3\textwidth]{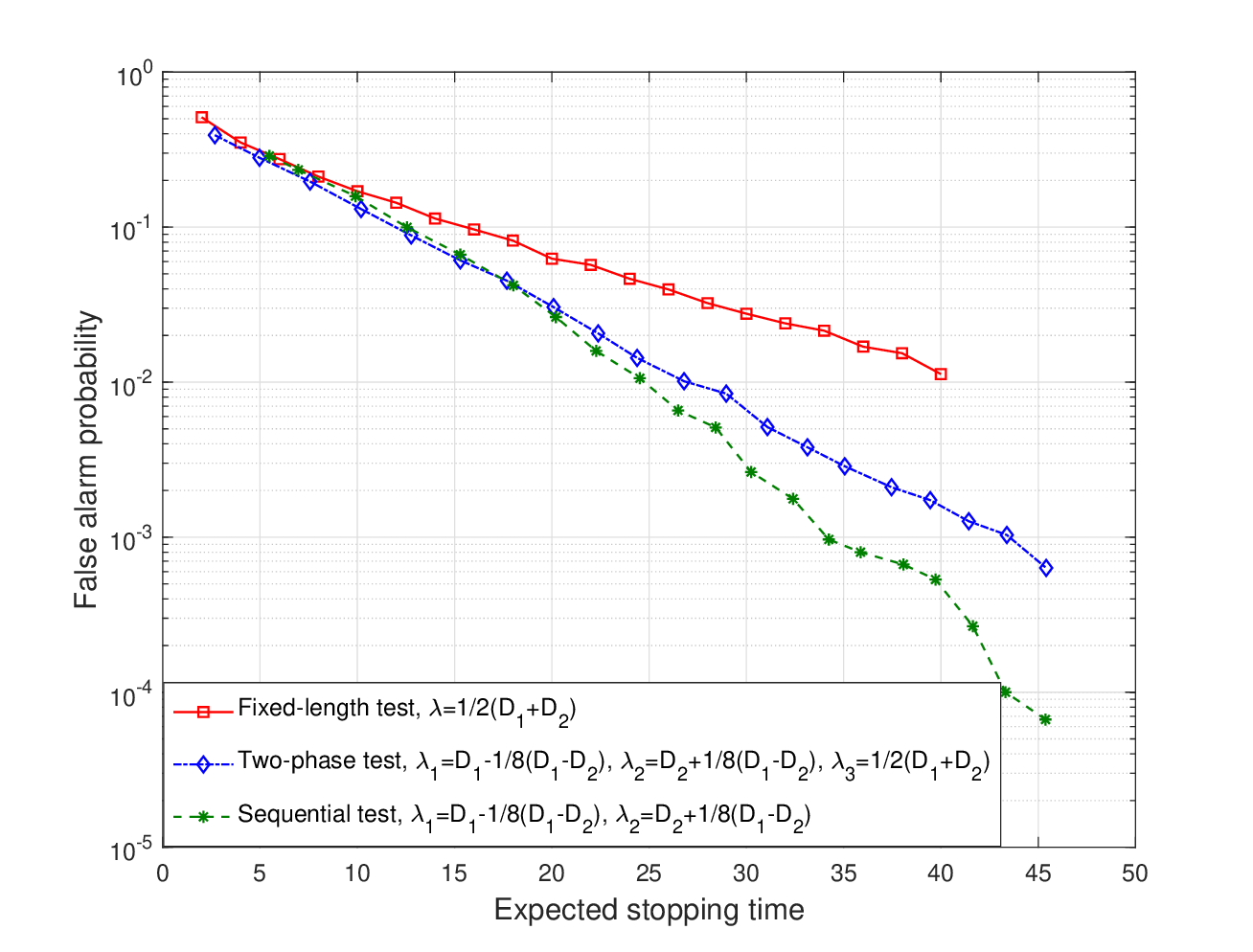}
}
\caption{Plot of simulated misclassification probabilities for our fixed-length test in Section \ref{S-FLMT-un}, sequential test in Section \ref{S-ST-un} and two-phase test in Section \ref{S-AFLMT_un}, under (a) hypothesis $\rmH_1$ is true and (b): null hypothesis. As observed, sequential and two-phase tests in Section \ref{S-ST-un}-\ref{S-AFLMT_un} outperform the fixed-length test in \ref{FLMT_un} as the expected stopping time $\bbE[\tau]$ tends to infinity.}
\label{detection_error_unknown}
\end{figure}

In Fig. \ref{detection_error_unknown}-(a), we plot the simulated misclassification probabilities of three tests in Section \ref{Main_un} under the non-null hypothesis. Furthermore, Fig. \ref{detection_error_unknown}-(b) plots the simulated false alarm error probabilities of our tests under the null hypothesis. Similar conclusions as in the simple case hold. In particular, our sequential and two-phase tests in Section \ref{S-ST-un}-Section \ref{S-AFLMT_un} outperform the fixed-length test in Section \ref{S-FLMT-un}.

\begin{figure}[htbp]
\centering
\includegraphics[height=0.3\textwidth]{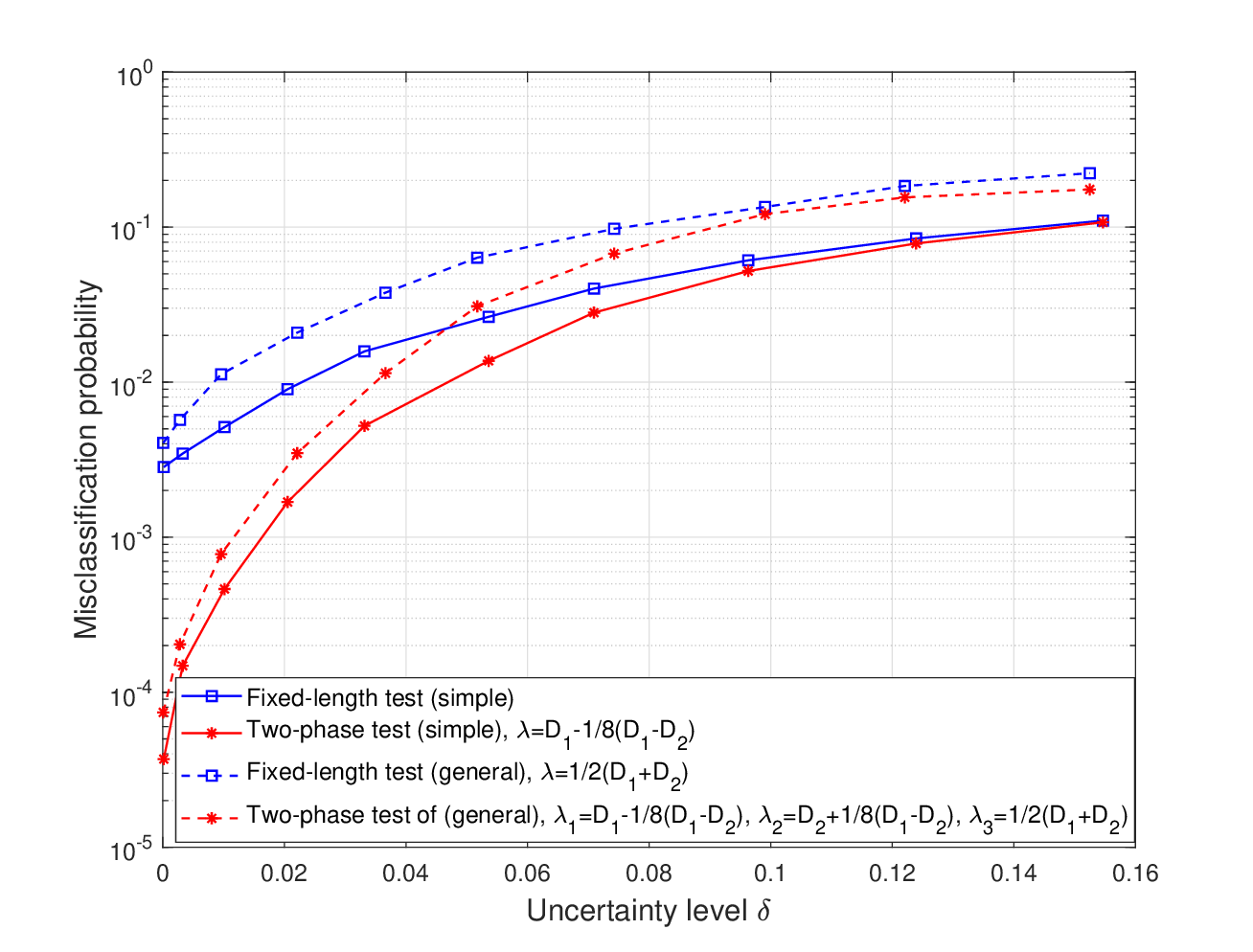}
\caption{Plot of simulated misclassification probabilities for our fixed-length and two-phase tests in Section \ref{Main} and \ref{Main_un} with $M=10$ training sequences when hypothesis $\rmH_1$ is true. As observed, there is a penalty in the performance of not knowing whether the null hypothesis is true.}
\label{detection_error_delta}
\end{figure}

Finally, in Fig. \ref{detection_error_delta}, for the same setting as Fig. \ref{detection_error_known} and Fig. \ref{detection_error_unknown}-(a), we compare performance of our fixed-length and two-phase tests with the varying of the uncertainty parameter $\delta$ and choosing the distance function $d(\cdot)$ to be the MMD function. It follows from the definition of $\calS_\delta(\cdot)$ in \eqref{MMDDistenceSet} that $\max_{Q\in \calS_\delta(P_i)}\mathrm{MMD}^2(Q,P_i)=\delta$. Let $n=25$ be the sampling length of the fixed-length test and that of the the two-phase test during the first phase.
As observed, when the uncertainty level $\delta$ increases, the misclassification probabilities of both tests increase. Furthermore, comparing the results of simple and general cases, we observe the penalty of not knowing whether the null hypothesis is true. On the other hand, as $\delta$ tends to zero, the penalty is negligible and the tests in Section \ref{Main_un} for the general case achieve roughly the same performance as the tests in the sample case.

\section{Conclusion}
\label{sec:conc}

We studied $M$-array classification for continuous sequences with distribution uncertainty, designed and characterized the achievable large deviations performance for three kinds of test using the MMD metric: fixed-length, sequential and two-phase tests. We considered both the simple case and the general case, where in the latter case, the testing sequence can be generated from a distribution that is vastly different from the generating distribution of any training sequence. In both cases, we showed that the two-phase test bridges over fixed-length and sequential tests by having performance close to the sequential test and having design complexity propositional to the fixed-length test. Our analyses in the general case involved a careful analysis of the false reject and false alarm error events concerning whether the testing sequence is generated from a distribution close to the generating distribution of one training sequence. Finally, we clarified the penalty of not knowing whether the null hypothesis is true.

We next discuss future directions. Firstly, we only derived achievable performance. It would be worthwhile to derive converse results to check whether our results are optimal in any regime. Towards this goal, one might modify the generalized Neyman-Pearson criterion used in the discrete case~\cite{second}. Secondly, we were interested in the asymptotic performance and proposed tests that have exponential complexity with respect to the number of outliers. For practical uses, it would be of interest to propose low-complexity tests that could achieve performance close to the benchmarks in this paper. To do so, one might generalize the clustering-based low-complexity test tailored for discrete sequences to continuous sequences~\cite{bu2019linear}. Furthermore, we assumed that each nominal sample follows the same unknown nominal distribution and each outlier follows the same unknown anomalous distribution. In practical applications, the nominal samples and outliers could have different distributions that center around certain distributions. For this case, it would be interesting to generalize the idea in \cite{binaryI-Hsiang,pan2022asymptotics} to outlier hypothesis testing of continuous sequences. Finally, it would be of value to generalize the ideas in this paper to other statistical inference problems, e.g., distributed detection \cite{tenney1981detection,tsitsiklis1988decentralized}, quickest change-point detection \cite{poor2008quickest,tartakovsky2014sequential} and clustering \cite{kaufman1990finding,park2009simple}.

\appendix
\subsection{Proof of Theorem \ref{FLMT}}
\label{proof_of_FLMT}
Consistent with \cite{MMD}, we need the following McDiarmid's inequality \cite{mcdiarmid1989method} to bound error probabilities.
\begin{lemma}
\label{McDiarmid}
Let $g:\calX^n\rightarrow \mathbb{R}$ be a
function such that for each $k\in[n]$, there exists a constant $c_k<\infty$ such that
\begin{align}\label{McDiarmid1}
\nn\sup_{x^n\in\calX^n,\tilx\in\calX}&\big|g(x_1,\ldots,x_k,\ldots,x_n)\\
&-g(x_1,\ldots, x_{k-1},\tilx,x_{k+1},\ldots,x_n)\big|\leq c_k.
\end{align}
Let $X^n$ be generated i.i.d. from a pdf $f\in\calP(\bbR)$. For any $\varepsilon>0$, it follows that
\begin{align}
\label{McDiarmid2}
\mathrm{Pr}\big\{g(X^n)-\bbE_f[g(X^n)]>\varepsilon\big\}< \exp\left\{-\frac{2\varepsilon^2}{\sum_{i=1}^nc_i^2}\right\}.
\end{align}
\end{lemma}

Recall that $\bY^N$ collects all training sequences $\{Y_1^N,\ldots,Y_M^N\}$, and $X^n$ is the testing sequence that is generated from an unknown distribution $Q$, where $Q\in\calS_\delta(P_i)$.
Given above definitions and recalling the definition of the test in \eqref{FLTest}, for each $i\in[M]$, we can bound the misclassification probability  $\beta_i(\phi_n|\bP,\delta)$ as follows:
\begin{align}
\nn&\beta_i(\phi_n|\bP,\delta)\\
&=\bbP_i\{\phi_n(X^n,\bY^N)\neq\rmH_i\}\label{beta_error_fx0}\\
&=\bbP_i\left\{i^*(X^n,\bY^N)\neq i\right\}\label{beta_error_fx1}\\
&\leq \bbP_i\left\{\exists~j\in \calM_i,\mathrm{MMD}^2(X^n,Y^N_i)>\mathrm{MMD}^2(X^n,Y^N_j)\right\}\\
&\leq\sum_{j\in\calM_i}\bbP_i\left\{\mathrm{MMD}^2(X^n,Y^N_i)>\mathrm{MMD}^2(X^n,Y^N_j)\right\}\label{beta_error_fx2}.
\end{align}

We apply McDiarmid's inequality to further upper bound \eqref{beta_error_fx2}. To do so, we need  to calculate the expected value of $\mathrm{MMD}^2(X^n,Y^N_i)-\mathrm{MMD}^2(X^n,Y^N_j)$ and the parameters $c_k$ for each $k\in[n+2N]$. Note that under hypothesis $\rmH_i$, $X^n$, $Y^N_i$, $Y^N_j$ are generated i.i.d. from the distributions $Q$, $P_i$ and $P_j$, respectively, where $Q\in \calS_{\delta}(P_i)$ and $Q\notin \calS_{\delta}(P_j)$. It follows from the definition of the MMD metric in \eqref{MMDcompute} that
\begin{align}
\nn&\bbE_{\bbP_i}[\mathrm{MMD}^2(X^n,Y^N_i)-\mathrm{MMD}^2(X^n,Y^N_j)]\\
&\leq D_2(\delta,M)-D_1(\delta,M).
\end{align}
To bound the Lipschitz constant $\{c_i\}$, given any testing sequence $x^n$ and training sequence $y^N_j$, define $g_{i,j}(x^n,y^N_i,y^N_j):=\mathrm{MMD}^2(x^n,y^N_i)-\mathrm{MMD}^2(x^n,y^N_j)$, which is a function of $n+2N$ parameters. For each $k\in[n+2N]$, if the $k$-th element of $\{x^n,y^N_i,y^N_j\}$ is replaced by $\tily$, we use $g_{i,j}(x^n,y^N_i,y^N_j,k,\tily)$ to denote the corresponding function value. For each $(j,k)\in\calM_i\times[n+2N]$, define
\begin{align}
c_k^{i,j}:=\sup_{x^n,y^N_i,y^N_j,\tily}|g_{i,j}(x^n,y^N_i,y^N_j)-g_{i,j}(x^n,y^N_i,y^N_j,k,\tily)|.
\end{align}

Given any $i\in[M]$, for each $j\in\calM_i$, and any $k\in[n+2N]$, similarly to [Eg. (27)-(32)]in \cite{MMD} and \cite{li2018nonparametric}, it follows that
\begin{align}
|c_k^{i,j}|&\leq \frac{8K_0}{n},~k\in[n],\\
|c_k^{i,j}|&\leq \frac{8K_0}{N},k\in[n+1,n+2N].
\end{align}
Thus,
\begin{align}\label{frac}
\sum_{k\in[n+2N]} (c_k^{i,j})^2\leq \frac{64K_0^2}{n}\left(1+\frac{2n}{N}\right)=\frac{64K_0^2}{n}\left(1+\frac{2}{\alpha}\right).
\end{align}

Using McDiarmid's inequality, with the definition of $g_1(\cdot)$ in \eqref{g1}, the misclassification probability in \eqref{beta_error_fx2} leads to
\begin{align}
\beta_i(\phi_n|\bP,\delta)
&\leq (M-1)\exp\left\{-ng_1(D_1,D_2)\right\},\label{beta_error_fx}
\end{align}
which is exponentially consistent.
The misclassification exponent is given by
\begin{align}\label{expbeta_fix}
& \lim_{n\to\infty} -\frac{1}{n}\log\beta_i(\phi_n|\bP,\delta)
\geq g_1(D_1,D_2).
\end{align}

\subsection{Proof of Theorem \ref{ST}}
\label{proof_of_ST}
To prove the theorem, we first bound the expected stopping time under each non-null hypothesis. Then we derive the error probability by following the similar idea of the analyses for the fixed-length test. Finally, the desired bound on the misclassification exponent can be obtained by combining the above analyses.

Recall the definition of the stopping time $\tau$ in \eqref{Taulength}. For each $i\in[M]$, the expected stopping time $\bbE_{\bbP_i}[\tau]$ can be upper bounded as follows:
\begin{align}
\bbE_{\bbP_i}[\tau]
&\leq \sum_{t=1}^\infty \bbP_i\{\tau\geq t\}\\
&\leq N_0-1+ \sum_{t=N_0-1}^\infty \bbP_i\{\tau>t\}\label{Etau},
\end{align}
where \eqref{Etau} follows since the random stopping time $\tau\ge N_0-1$ by the definition in \eqref{Taulength}.

If $\lambda<D_1(\delta,M)$, each probability term inside the sum of \eqref{Etau} satisfies
\begin{align}
\nn&\bbP_i\{\tau>t\}\\
&=\bbP_i\{h(X^t,\bY^{\alpha t})<\lambda\}\\
\nn&\leq\bbP_i\Big\{\exists~(j,k)\in\calM^2,~j\neq k:~\mathrm{MMD}^2(X^t,Y^{\alpha t}_j)<\lambda,\\
&\qquad\qquad \mathrm{MMD}^2(X^t,Y^{\alpha t}_k)<\lambda\Big\}\label{Pbetatau1}\\
\nn&\leq \sum_{j\in[M]}\sum_{k\in\calM_j}\bbP_i\{\mathrm{MMD}^2(X^t,Y^{\alpha t}_j)<\lambda\mathrm{~and~}\\
&\qquad\qquad\qquad\qquad\mathrm{MMD}^2(X^t,Y^{\alpha t}_k)<\lambda\}\\
\nn&\leq \sum_{j\in\calM_i}\sum_{k\in\calM_j}\bbP_i\{\mathrm{MMD}^2(X^t,Y^{\alpha t}_j)<\lambda\}\\
&+\sum_{k\in\calM_i}\bbP_i\{\mathrm{MMD}^2(X^t,Y^{\alpha t}_k)<\lambda\}\label{Pbetatau2}\\
&\leq  M\sum_{k\in\calM_i}\bbP_i\{\mathrm{MMD}^2(X^t,Y^{\alpha t}_k)<\lambda\}\label{Pbetatau3}.
\end{align}
To further upper bound \eqref{Pbetatau3}, we follow the similar idea as in the fixed-length test by applying McDiarmid's inequality.
Define the function $g_{i,j}(x^t,y^{\alpha t}_j):=\mathrm{MMD}^2(x^t,y^{\alpha t}_{j})$. For each $k\in[t+\alpha t]$, if the $k$-th element of $\{x^t,y^{\alpha t}_j\}$ is replaced by $\tily$, we use $g_{i,j}(x^t,y^{\alpha t}_j,k,\tily)$ to denote the corresponding function value. For each $(j,k)\in\calM_i\times[t+\alpha t]$, define
\begin{align}
c_k^{i,j}:=\sup_{x^t,y^{\alpha t}_j,\tily}|g_{i,j}(x^t,y^{\alpha t}_j)-g_{i,j}(x^t,y^{\alpha t}_j,k,\tily)|.
\end{align}
It follows from the definition of the MMD metric in \eqref{MMDcompute} that
\begin{align}
&\bbE_{\bbP_i}[\mathrm{MMD}^2(X^t,Y^{\alpha t}_j)]=\mathrm{MMD}^2(Q,P_j)\geq D_1(\delta,M).\\
&\sum_{k\in[t+{\alpha t}]} (c_k^{i,j})^2\leq \frac{64K_0^2}{t}\left(1+\frac{1}{\alpha}\right).\label{frac2}
\end{align}
With the results above and the definition of $g_2(\cdot)$ in \eqref{g2}, we can upper bound \eqref{Pbetatau3} by
\begin{align}
\bbP_i\{\tau>t\}
&\leq(M-1)M\exp\left\{-tg_2(\lambda)\right\},\label{Ptau}
\end{align}
Combining \eqref{Etau} and \eqref{Ptau} leads to
\begin{align}
\bbE_{\bbP_i}[\tau]&\leq N_0-1+\sum_{t=N_0-1}^\infty (M-1)M\exp\left\{-tg_2(\lambda)\right\}\label{sumEtau}.
\end{align}
When $N_0$ is sufficiently large, it follows from \eqref{sumEtau} that $\bbE_{\bbP_i}[\tau]\leq N_0$ when $\lambda<D_1(\delta,M)$. On the other hand, if $\lambda\geq D_1(\delta,M)$, $\bbP_i\{\tau>t\}\leq 1$ and $\bbE_{\bbP_i}[\tau]\leq \infty$.

For each $i\in[M]$, the misclassification probability $\beta_i^{\mathrm{seq}}(\phi_\tau|\bP,\delta)$ satisfies
\begin{align}
\nn&\beta_i^{\mathrm{seq}}(\phi_\tau|\bP,\delta)\\
&\leq \sum_{t=N_0-1}^\infty\bbP_i\{\phi_t(X^t,\bY^{\alpha t})\neq\rmH_i\}\\
&\leq \sum_{t=N_0-1}^\infty\bbP_i\{i^*(X^t,\bY^{\alpha t})\neq i, 
\text{ and } h(X^t,\bY^{\alpha t})>\lambda\}\\
\nn&\leq \sum_{t=N_0-1}^\infty\min\big\{\bbP_i\{i^*(X^t,\bY^{\alpha t})\neq i\},\\ 
&\qquad\qquad\qquad\bbP_i\{\mathrm{MMD}^2(X^t,Y^{\alpha t}_i)>\lambda\}\big\}\label{Pbetas0}.
\end{align}
The first item in the minimization of \eqref{Pbetas0} can be easily obtained by following
\eqref{beta_error_fx} with $n$ replaced with $t$, i.e.,
\begin{align}
\bbP_i\{i^*(X^t,\bY^{\alpha t})\neq i\}\leq (M-1)\exp\left\{-tg_1(D_1,D_2)\right\}.\label{Pbetas1}
\end{align}

We then apply McDiarmid's inequality to further upper bound the second item in the minimization of  \eqref{Pbetas0}.
It follows from the definition of MMD metric in \eqref{MMDcompute} that
\begin{align}
&\bbE_{\bbP_i}[\mathrm{MMD}^2(X^t,Y^{\alpha t}_i)]=\mathrm{MMD}^2(Q,P_i)\leq D_2(\delta,M).
\end{align}
Using McDiarmid's inequality and by following the similar idea of \eqref{frac}, if $\lambda>D_2(\delta,M)$, one can obtain
\begin{align}
\bbP_i\{\mathrm{MMD}^2(X^t,Y^{\alpha t}_i)>\lambda\}
&\leq \exp\left\{-tg_3(\lambda)\right\},\label{Pbetas2}
\end{align}
where $g_3(\cdot)$ was defined in \eqref{g3}.

Using \eqref{sumEtau}, \eqref{Pbetas0}, \eqref{Pbetas1}, and \eqref{Pbetas2}, when $D_2(\delta,M)<\lambda<D_1(\delta,M)$, we conclude that 
\begin{align}
\nn&\beta_i^{\mathrm{seq}}(\phi_\tau|\bP,\delta)
\leq \\
&\sum_{t=N_0-1}^\infty\min\Big\{(M-1)\exp\left\{-tg_1(D_1,D_2)\right\},
\exp\left\{-tg_3(\lambda)\right\}\Big\}\label{beta_error_ST}
\end{align}
and the misclassification exponent satisfies
\begin{align}
&\lim_{N_0\rightarrow\infty}-\frac{1}{\bbE_{\bbP_i}[\tau]}\log \beta_i^{\mathrm{seq}}(\phi_\tau|\bP,\delta)\geq \max\Big\{g_1(D_1,D_2),g_3(\lambda)\Big\}.\label{logbetas}
\end{align}
Otherwise, if $\lambda<D_2(\delta,M)$, $h(X^t,\bY^{\alpha t})>\lambda$ can always be satisfied and the test will stop at the beginning of $\tau=N_0-1$. The error exponent is the same as that of fixed-length test. 

\subsection{Proof of Theorem \ref{FLMT_un}}
\label{proof_of_FLMT_Un}
Recall the definitions of  $\Bar{g}_1(\cdot)$, $\Bar{g}_2(\cdot)$,  and $\Bar{g}_3(\cdot)$.
Given the test in \eqref{FLTest_un}, we can bound the misclassification and false alarm exponents.  For each $i\in[M]$, when $\lambda<\Bar{D}_1(\delta,M)$ the misclassification probability $\beta_i(\phi_n|\bP,\delta)$ satisfies
\begin{align}
\beta_i(\phi_n|\bP,\delta)
&=\bbP_i\{\phi_n(X^n,\bY^N)\neq\rmH_i\}\\
\nn&=\bbP_i\{\phi_n(X^n,\bY^N)\notin\{\rmH_i,\rmH_\rmr\}\}\\*
&\qquad+\bbP_i\{\phi_n(X^n,\bY^N)=\rmH_\rmr\}\label{betaun_error_fx0}
\end{align}
For the first item in \eqref{betaun_error_fx0}, when $\lambda<\Bar{D}_1(\delta,M)$, we have
\begin{align}
\nn&\bbP_i\{\phi_n(X^n,\bY^N)\notin\{\rmH_i,\rmH_\rmr\}\}\\
&=\bbP_i\{\exists~j\in\calM_i,i^*(X^n,\bY^N)=j, \mathrm{MMD}^2(X^n,Y^n_j)<\lambda\}\label{betaun_error_fx1}\\
\nn&\leq \bbP_i\big\{ \exists~j\in\calM_i, \mathrm{MMD}^2(X^n,Y^N_i)>\mathrm{MMD}^2(X^n,Y^N_j), \\
&\qquad\qquad\mathrm{MMD}^2(X^n,Y^N_j)<\lambda\big\}\\
\nn&\leq \sum_{j\in\calM_i}\bbP_i\big\{\mathrm{MMD}^2(X^n,Y^N_i)>\mathrm{MMD}^2(X^n,Y^N_j), \\
&\qquad\qquad\mathrm{MMD}^2(X^n,Y^N_j)<\lambda\big\}\label{betaun_error_fx2}\\
\nn&\leq \sum_{j\in\calM_i}\min\big\{\bbP_i\{\mathrm{MMD}^2(X^n,Y^N_i)>\mathrm{MMD}^2(X^n,Y^N_j)\},\\
&\qquad\qquad\bbP_i\left\{\mathrm{MMD}^2(X^n,Y^N_j)<\lambda\right\}\big\}\\
&\leq (M-1)\min\big\{\exp\{-n\Bar{g}_1(\Bar{D}_1,\Bar{D}_2)\},\exp\{-n\Bar{g}_2(\lambda)\}\big\}\label{betaun_error_fx},
\end{align}
where \eqref{betaun_error_fx} follows the results of \eqref{beta_error_fx}, \eqref{Pbetatau3} and \eqref{Ptau}.

Then we consider the second first item in \eqref{betaun_error_fx0}.
For each $i\in[M]$, when $\lambda> \Bar{D}_2(\delta,M)$, we have
\begin{align}
\nn&\bbP_i\{\phi_n(X^n,\bY^N)=\rmH_\rmr\}\\
&=\bbP_i(\forall~j\in[M],\mathrm{MMD}^2(X^n,Y^N_j)>\lambda)\label{zetaun_error_fx0}\\
& \leq \bbP_i\{\mathrm{MMD}^2(X^n,Y^N_i)>\lambda\}\\
&\leq \exp\left\{-n\Bar{g}_3(\lambda)\right\},\label{zetaun_error_fx}
\end{align}
where \eqref{zetaun_error_fx} follows the result of \eqref{Pbetas2}.
Combing \eqref{betaun_error_fx} and \eqref{zetaun_error_fx}, when $\Bar{D}_2(\delta,M)<\lambda<\Bar{D}_1(\delta,M)$, we can obtain the 
 misclassification exponent as follows:
\begin{align}\label{expzeta_fix_un}
& \lim_{n\to\infty} -\frac{1}{n}\log\beta_i(\phi_n|\bP,\delta)\\
& \geq  \min\big\{\max\{\Bar{g}_1(\Bar{D}_1,\Bar{D}_2),\Bar{g}_2(\lambda)\},\Bar{g}_3(\lambda)\big\}.
\end{align}
On the other hand, if $\lambda\geq \Bar{D}_1(\delta,M)$, the misclassification exponent has the same form as that of fixed-length test in the simple case. If $\lambda\leq \Bar{D}_2(\delta,M)$, the exponent tends to zero.

Analogously, for the case of null hypothesis, $X^n$ is generated from an unknown distribution $Q$ which satisfies $Q\notin\cup_{i=1}^M\calS_\delta(P_i)$. When $\lambda<\Bar{D}_1(\delta,M)$
The false alarm probability $\rmP_{\mathrm{FA}}$ is upper
bounded as follows
\begin{align}
\rmP_{\mathrm{FA}}(\phi_n|\bP,\delta)
&=\bbP_\rmr\{\phi_n(X^n,\bY^N)\neq \rmH_\rmr\}\\
&=\bbP_\rmr\{\exists~j\in[M],i^*(X^n,\bY^N)=j, \\
&\qquad\qquad\text{ and }\mathrm{MMD}^2(X^n,Y^n_j)<\lambda\}\\
&\leq \bbP_\rmr\{\exists~j\in[M], \mathrm{MMD}^2(X^n,Y^N_j)<\lambda\}\\
&\leq\sum_{j\in[M]} \bbP_\rmr\left\{\mathrm{MMD}^2(X^n,Y^N_j)<\lambda\right\}\\
&\leq M\exp\left\{-n\Bar{g}_2(\lambda)\right\},\label{Faun_error_fx}
\end{align}
where \eqref{Faun_error_fx} follows the results of \eqref{Pbetatau3} and \eqref{Ptau}. The false alarm exponent is given by
\begin{align}\label{expFA_fix_un}
& \lim_{n\to\infty} -\frac{1}{n}\log\rmP_{\mathrm{FA}}(\phi_n|\bP,\delta)
\geq  \Bar{g}_2(\lambda).
\end{align}
If $\lambda\geq \Bar{D}_1(\delta,M)$, the exponent tends to zero.

\subsection{Proof of Theorem \ref{ST_un}}
\label{proof_of_ST_un}
To prove the theorem, analogously to the analyses for the sequential test in the simple case, we need to bound the expected stopping time and the error exponents under each non-null and null hypothesis. 

Recall the definition of the stopping time $\tau$ in \eqref{Taulength_un}. For each $i\in[M]$, the expected stopping time $\bbE_{\bbP_i}[\tau]$ can be upper bounded as follows:
\begin{align}
\bbE_{\bbP_i}[\tau]
&\leq \sum_{t=1}^\infty \bbP_i\{\tau\geq t\}\\
&\leq N_0-1+ \sum_{t=N_0-1}^\infty \bbP_i\{\tau>t\}\label{Etau_un}.
\end{align}
If $\lambda_2<\Bar{D}_1(\delta,M)$ and $\lambda_1>\Bar{D}_2(\delta,M)$, each probability term inside the sum of \eqref{Etau_un} satisfies
\begin{align}
\nn&\bbP_i\{\tau>t\}\\
&=\bbP_i\{h(X^t,\bY^{\alpha t})<\lambda_2\}\\
&+\bbP_i\{\lambda_1<\min_{j\in[M]}\mathrm{MMD}^2(X^t,Y_j^{\alpha t})<\lambda_2\}\\
&\leq \bbP_i\{h(X^t,\bY^{\alpha t})<\lambda_2\}+\bbP_i\{\min_{j\in[M]}\mathrm{MMD}^2(X^t,Y_j^{\alpha t})>\lambda_1\}\\
&\leq \bbP_i\{h(X^t,\bY^{\alpha t})<\lambda_2\}+\bbP_i\{\mathrm{MMD}^2(X^t,Y_i^{\alpha t})>\lambda_1\}\\
&\leq (M-1)M\exp\left\{-t\Bar{g}_2(\lambda_2)\right\}+\exp\left\{-t\Bar{g}_3(\lambda_1)\right\},\label{Ptau_un}
\end{align}
where \eqref{Ptau_un} follows similarly to \eqref{Ptau} and \eqref{Pbetas2} except that $\lambda$ is replaced with $\lambda_2$ and $\lambda_1$.
Combining \eqref{Etau_un} and \eqref{Ptau_un} leads to
\begin{align}
\nn&\bbE_{\bbP_i}[\tau]\leq N_0-1\\
&+\sum_{t=N_0-1}^\infty (M-1)M\exp\left\{-t\Bar{g}_2(\lambda_2)\right\}+\exp\left\{-t\Bar{g}_3(\lambda_1)\right\}.\label{sumEtau_un}
\end{align}
If $\lambda_2\geq \Bar{D}_1(\delta,M)$ or $\lambda_1\leq \Bar{D}_2(\delta,M)$, $\bbP_i\{\tau>t\}\leq 1$ and $\bbE_{\bbP_i}[\tau]\leq \infty$.

Similarly, when $\lambda_2<\Bar{D}_1(\delta,M)$, the expected stopping time under the null hypothesis satisfies:
\begin{align}
&\bbE_{\bbP_\rmr}[\tau]
\leq  N_0-1+ \sum_{t=N_0-1}^\infty \bbP_\rmr\{\tau> t\}\\
&\leq N_0-1+\sum_{t=N_0-1}^\infty\bbP_\rmr\{\min_{j\in[M]}\mathrm{MMD}^2(x^t,y_j^{\alpha t})<\lambda_2\}\\
&\leq N_0-1+\sum_{t=N_0-1}^\infty M\exp\left\{-t\Bar{g}_2(\lambda_2)\right\},\label{Etau2_un}
\end{align}
where \eqref{Etau2_un} follows the results in \eqref{Faun_error_fx} by replacing $\lambda$ with $\lambda_2$.
On the other hand, if $\lambda_2\geq \Bar{D}_1(\delta,M)$, $\bbP_\rmr\{\tau>t\}\leq 1$ and $\bbE_{\bbP_\rmr}[\tau]\leq \infty$.

When $N_0$ is sufficiently large, it follows from \eqref{sumEtau_un} and \eqref{Etau2_un} that $\bbE_{\bbP_i}[\tau]\leq N_0$ and $\bbE_{\bbP_\rmr}[\tau]\leq N_0$. For each $i\in[M]$, when $\Bar{D}_2(\delta,M)<\lambda_1<\lambda_2<\Bar{D}_1(\delta,M)$, the misclassification probability $\beta_i^{\mathrm{seq}}(\phi_\tau|\bP,\delta)$ satisfies
\begin{align}
\beta_i^{\mathrm{seq}}(\phi_\tau|\bP,\delta)
&\leq \sum_{t=N_0-1}^\infty\bbP_i\{\phi_t(X^t,\bY^{\alpha t})\neq\rmH_i\}\\
\nn&\leq \sum_{t=N_0-1}^\infty\bbP_i\{\phi_t(X^t,\bY^{\alpha t})\notin\{\rmH_i,\rmH_\rmr\}\}\\
&\qquad+\sum_{t=N_0-1}^\infty\bbP_i\{\phi_t(X^t,\bY^{\alpha t})=\rmH_\rmr\}\label{betaS_un0}.
\end{align}
The first item in \eqref{betaS_un0} can be rewritten by
\begin{align}
&\bbP_i\{\phi_t(X^t,\bY^{\alpha t})\notin\{\rmH_i,\rmH_\rmr\}\}\\
\nn&\leq \bbP_i\{\exists~j\in\calM_i,i^*(X^t,\bY^{\alpha t})=j, \mathrm{MMD}^2(X^t,Y^{\alpha t}_j)<\lambda_1\\
&\qquad\qquad\text{ and } h(X^t,\bY^{\alpha t})>\lambda_2\}\\
\nn&\leq \min\big\{\bbP_i\{\exists~j\in\calM_i,i^*(X^t,\bY^{\alpha t})=j, \\
\nn&\qquad\qquad\text{ and }\mathrm{MMD}^2(X^t,Y^{\alpha t}_j)<\lambda_1\},\\
&\qquad\qquad\bbP_i\{\mathrm{MMD}^2(X^t,Y^{\alpha t}_i)>\lambda_2\}\big\}\\
\nn&\leq\min\big\{
(M-1)\exp\left\{-t\Bar{g}_1(\Bar{D}_1,\Bar{D}_2)\right\},\\
&\qquad\quad(M-1)\exp\big\{-t\Bar{g}_2(\lambda_1)\big\},\exp\big\{-t\Bar{g}_3(\lambda_2)\big\}
\big\}\label{betaun_error_ST},
\end{align}
where \eqref{betaun_error_ST} follows the results in \eqref{betaun_error_fx} and \eqref{Pbetas2}.

Similarly, the second item in \eqref{betaS_un0} is upper bounded by:
\begin{align}
\nn&\bbP_i\left\{\phi_t(X^t,\bY^{\alpha t})=\rmH_\rmr\right\}\\
&\leq \bbP_i\left\{\forall~j\in[M],\mathrm{MMD}^2(X^t,Y^{\alpha t}_j)>\lambda_2\right\}\\
&\leq \exp\left\{-t\Bar{g}_3(\lambda_2)\right\}.\label{zetaun_error_ST}
\end{align}
where \eqref{zetaun_error_ST} follows the result of \eqref{zetaun_error_fx} with $(n,\lambda)$ replaced with $(t,\lambda_2)$. 

With the results of \eqref{betaun_error_ST} and \eqref{zetaun_error_ST}
The error exponent of the misclassification probability is given by
\begin{align}
\lim_{N_0\rightarrow\infty}-\frac{1}{\bbE_{\bbP_i}[\tau]}\log \beta_i^{\mathrm{seq}}(\phi_\tau|\bP,\delta)
&\geq \Bar{g}_3(\lambda_2)\label{logbetas_un},
\end{align}

Finally, when $\lambda_1<\Bar{D}_1(\delta,M)$, the false alarm probability $\rmP_{\mathrm{FA}}^{\mathrm{seq}}(\phi_\tau|\bP,\delta)$ is upper bounded by
\begin{align}
\nn&\rmP_{\mathrm{FA}}^{\mathrm{seq}}(\phi_\tau|\bP,\delta)\\
&\leq \sum_{t=N_0-1}^\infty
\bbP_\rmr\left\{\phi_t(X^t,\bY^{\alpha t})\neq\rmH_\rmr\right\}\\
\nn&= \sum_{t=N_0-1}^\infty
\bbP_\rmr\big\{\exists~j\in[M],i^*(X^t,\bY^{\alpha t})=j, \\
&\qquad\mathrm{MMD}^2(X^t,Y^{\alpha t}_j)<\lambda_1\text{ and } h(X^t,\bY^{\alpha t})>\lambda_2\big\}\\
&\leq \sum_{t=N_0-1}^\infty
\bbP_\rmr\left\{\exists~j\in[M],\mathrm{MMD}^2(X^t,Y^{\alpha t}_j)<\lambda_1\right\}\\
&\leq \sum_{t=N_0-1}^\infty M\exp\left\{-t\Bar{g}_2(\lambda_1)\right\}\label{FAun_error_ST},
\end{align}
where \eqref{FAun_error_ST} follows the result of \eqref{Faun_error_fx} with $(n,\lambda)$ replaced with $(t,\lambda_1)$.
The false alarm exponent satisfies
\begin{align}
\liminf_{N_0\to\infty}-\frac{1}{\bbE_{\bbP_\rmr}[\tau]}\log \rmP^{\rmseq}_{\rm{FA}}(\phi_\tau|\bP,\delta)
\geq \Bar{g}_2(\lambda_1).
\end{align}

\bibliographystyle{IEEEtran}
\bibliography{BiB}
\end{document}